\documentclass{article}

\PassOptionsToPackage{numbers,sort&compress}{natbib}

\usepackage[main,final]{neurips_2026}

\usepackage[utf8]{inputenc} 
\usepackage[T1]{fontenc}    
\usepackage{hyperref}       
\usepackage{url}            
\usepackage{booktabs}       
\usepackage{amsfonts}       
\usepackage{nicefrac}       
\usepackage{microtype}      
\usepackage{xcolor}         
\usepackage{amsmath}
\usepackage{graphicx}
\usepackage{multirow}
\usepackage{threeparttable}
\usepackage{makecell}
\usepackage[most]{tcolorbox}
\usepackage{enumitem}
\usepackage[american]{babel}
\usepackage{microtype}
\usepackage{xcolor}
\newcommand{\eg}{\emph{e.g.}}
\newcommand{\ie}{\emph{i.e.}}
\usepackage{tabularx}

\title{Learning to Group and Order: Cross-Instance Self-Supervised RL for Vision-Centric MLLMs}

\author{%
  Songming Yang\textsuperscript{1} \qquad
  Hong-Tao Yu\textsuperscript{1} \qquad
  Xiu-Shen Wei\textsuperscript{1,2}%
  \thanks{Corresponding author. This work was supported by National Natural Science Foundation of China under Grant (62522602, 62272231), Basic Research Program of Jiangsu under Grant (BK20250073), the Fundamental Research Funds for the Central Universities (4009002401, 2242025K30024), and the Big Data Computing Center of Southeast University.}\\
  \textsuperscript{1}School of Computer Science and Engineering, Southeast University \\
  \textsuperscript{2}Key Laboratory of New Generation Artificial Intelligence Technology \\
  and Its Interdisciplinary Applications, Southeast University \\
  \texttt{\{yangsm,yuht,weixs\}@seu.edu.cn}
}

\begin{document}

\maketitle

\begin{abstract}
Reinforcement learning (RL) post-training has become an important paradigm for improving multimodal large language models (MLLMs), especially with automatically verifiable rewards. Self-supervised visual pretext tasks provide such rewards without human annotations. However, existing methods mainly exploit within-instance structure, offering limited supervision for cross-instance discrimination and fine-grained comparison. We introduce \textbf{G}roup-and-\textbf{O}rder \textbf{S}elf-\textbf{S}upervised Reinforcement \textbf{L}earning (GO-SSL), a cross-instance jigsaw framework for vision-centric MLLM post-training. Given two visual instances, GO-SSL mixes their local elements and trains the model to group them by instance of origin while recovering the order within each group, coupling instance-level comparison with spatial, temporal, or geometric structure recovery. This paired formulation turns the same unlabeled data into richer verifiable supervision through diverse pairings, shuffles, and a random-to-hard curriculum. We instantiate GO-SSL mainly on cross-image jigsaw tasks and further extend it to 3D depth and video temporal ordering. Across diverse benchmarks, GO-SSL consistently improves over the base MLLM and prior self-supervised post-training baselines, with clear gains in fine-grained perception and spatial understanding. Further evaluations across model scales and families support its generality. These results suggest that cross-instance comparison provides a useful self-supervised RL signal for improving transferable visual understanding in MLLMs. Code is available at \url{https://github.com/SEU-VIPGroup/GO-SSL}.
\end{abstract}

\section{Introduction}
\label{sec:Introduction}
Multimodal large language models (MLLMs) have made remarkable progress in visual question answering and multimodal reasoning~\cite{yin2024survey}, as exemplified by LLaVA-NeXT~\cite{liu2024llavanext}, InternVL~\cite{chen2024internvl}, and Qwen2.5-VL~\cite{bai2025qwen25vl}. Nevertheless, their general-purpose capabilities do not always translate into reliable vision-centric understanding. Fine-grained recognition often requires distinguishing subtle local evidence~\cite{tong2024eyes,wu2024vstar,yu2026benchmarking,gs26_peng2026survey}, while spatial and relational tasks demand accurate reasoning about visual structure~\cite{liu2023vsr,jia2025omnispatial}. Improving these abilities calls for learning signals that directly target how models inspect and organize visual information.

Post-training is a promising way to strengthen such capabilities without redesigning the underlying MLLM. Supervised fine-tuning can improve visual behavior, but collecting and annotating task-specific examples across diverse visual domains is costly. Reinforcement learning with verifiable rewards (RLVR)~\cite{shao2024deepseekmath,guo2025deepseekr1} offers another route: a model can learn from feedback when the correctness of its output is automatically checkable. The challenge for vision-centric post-training is therefore to design tasks whose answers are verifiable while still requiring meaningful visual understanding.

Self-supervised learning (SSL) offers a natural solution to this challenge. Classical visual pretext tasks, including context prediction, rotation prediction, and jigsaw solving~\cite{doersch2015context,gidaris2018rotation,noroozi2016jigsaw}, derive supervision from input transformations rather than manual labels. Recent methods such as SSL4RL~\cite{guo2025ssl4rl}, Spatial-SSRL~\cite{liu2026spatialssrl}, and Visual Jigsaw~\cite{wu2026visualjigsaw} bring such tasks into RL post-training using verifiable rewards. However, jigsaw-style objectives largely recover spatial
or temporal structure within a single source.
They teach the model where an element belongs, but not
which visual instance it belongs to.
This leaves cross-instance discrimination without explicit
supervision, even when distinguishing similar instances
requires subtle local cues.

\begin{figure}[t]
  \centering
  \includegraphics[width=0.85\linewidth]{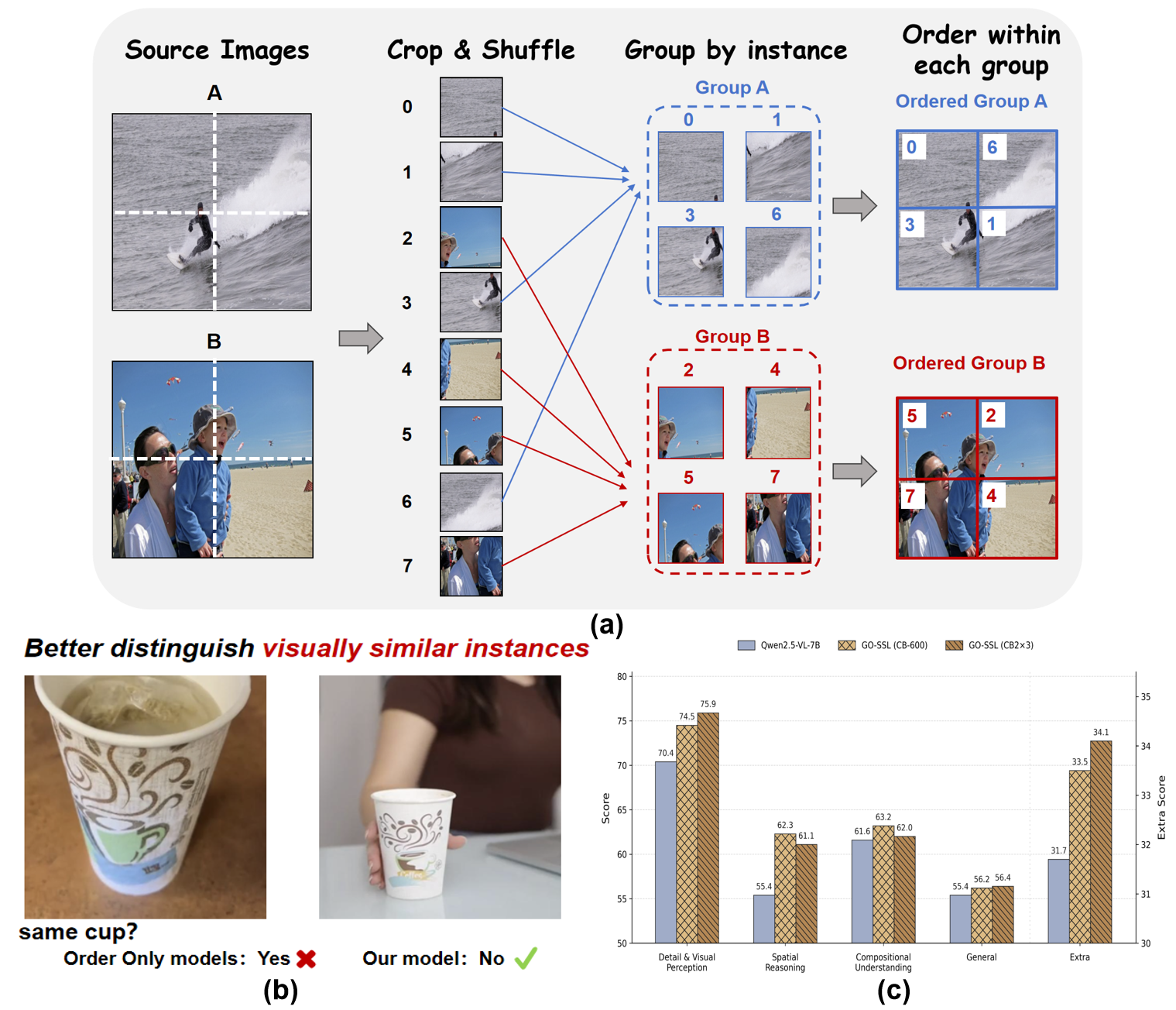}
  \caption{
    \small GO-SSL: task construction and visual transfer.
    \textbf{(a)} Shuffled patches from two images are grouped by source and
    reordered to recover each image's spatial layout.
    \textbf{(b)} A qualitative example of distinguishing visually similar
    object instances.
    \textbf{(c)} Category-wise benchmark averages for Qwen2.5-VL-7B and two
    GO-SSL variants, showing gains across all five categories.
    }
  \label{fig:intro}

\end{figure}

To make cross-instance comparison an explicit part of
self-supervised learning, we propose
\emph{Group-and-Order Self-Supervised Reinforcement Learning}
(GO-SSL). As shown in Fig.~\ref{fig:intro}(a), GO-SSL interleaves local elements from two sources and asks the model to recover both their source membership and their order within each source. Grouping makes cross-instance comparison an explicit requirement, whereas ordering preserves the need to reconstruct meaningful visual structure. Both targets are known from task construction and can therefore be checked automatically. Figure~\ref{fig:intro}(b) illustrates the type of fine-grained instance distinction that motivates this design.

Pairing also creates diverse comparative contexts from
the same unlabeled pool.
We exploit this flexibility through a random-to-hard
curriculum that makes source discrimination more challenging. We study image jigsaw reconstruction as the primary setting and extend the same principle to cross-scene depth ranking and cross-video temporal ordering. Across image, 3D, and video benchmarks, GO-SSL improves upon the base MLLM and prior self-supervised post-training baselines; Fig.~\ref{fig:intro}(c) summarizes the image-side gains. We additionally examine grouping and ordering ablations, model-family generalization, and transfer to targeted cross-instance discrimination tasks.

Our contributions are summarized as follows:
\begin{itemize}[leftmargin=1.0em, itemsep=0.15em, topsep=0.2em, parsep=0pt, partopsep=0pt]
    \item We propose GO-SSL, a cross-instance self-supervised RL framework that converts unlabeled visual data into verifiable group-and-order tasks, coupling fine-grained instance discrimination with spatial, temporal, and geometric structure recovery.
    \item We show that pairing visual instances creates diverse comparative contexts from the same unlabeled pool, with random-to-hard curricula controlling source ambiguity and task granularity.
    \item We demonstrate improvements over prior self-supervised post-training baselines on image, 3D, and video benchmarks, and further evaluate generalization across model scales and families and transfer to targeted cross-instance discrimination tasks.
\end{itemize}

\section{Related Work}
Our work connects self-supervised RL post-training with vision-centric understanding. We review these directions with an emphasis on verifiable visual supervision.

\paragraph{Self-supervised RL post-training for MLLMs}
Self-supervised learning (SSL) derives supervision from intrinsic data structure, with classical tasks including context prediction, rotation prediction, and jigsaw solving~\cite{doersch2015context,gidaris2018rotation,noroozi2016jigsaw}. Modern SSL employs masked reconstruction, contrastive learning, and self-distillation~\cite{bao2022beit,he2022mae,chen2020simclr,he2020moco,grill2020byol,caron2021dino}. Fine-grained approaches further explore progressive augmentation~\cite{gs26_tan2025prototype} and asymmetric augmentation for image hashing~\cite{gs26_hu2024asymmetric}; related detection pretraining combines holistic-local contrastive learning with reconstruction~\cite{gs26_duan2024longtailed}. Although effective for representation learning, these objectives are not directly aligned with MLLMs' discrete outputs.

Post-training and alignment methods, including RLHF, DPO, and RL with verifiable rewards (RLVR), have substantially improved large models~\cite{ouyang2022instructgpt,rafailov2023dpo,shao2024deepseekmath,guo2025deepseekr1}. Many multimodal methods still emphasize text-centric reasoning, where images serve mainly as supporting evidence. To provide more direct visual feedback, SSL4RL constructs intrinsic verifiable rewards from self-supervised tasks~\cite{guo2025ssl4rl}, and Visual Jigsaw demonstrates the benefits of ordering-based supervision across images, videos, and 3D inputs~\cite{wu2026visualjigsaw}. These studies show how visual pretext tasks can supply automatically checkable RL targets, while existing jigsaw-style objectives mainly use within-source structural signals. GO-SSL instead makes cross-instance source discrimination explicit alongside structural ordering.

\paragraph{Vision-centric understanding}
Fine-grained recognition depends on discriminative local and global visual cues~\cite{yang2026fgmoe}, yet fine-grained perception and spatial reasoning remain challenging for MLLMs. Benchmarks including ViewSpatial-Bench, Spatial457, and 3DSRBench highlight limitations in viewpoint-dependent and 3D reasoning~\cite{li2025viewspatial,wang2025spatial457,ma20253dsrbench}. Recent self-supervised methods address related capabilities through verifiable visual tasks. Spatial-SSRL~\cite{liu2026spatialssrl} constructs patch reordering, flip recognition, inpainting, depth ordering, and relative 3D position tasks, while Visual Jigsaw~\cite{wu2026visualjigsaw} recovers structural order across different visual domains.

Other related directions include object-level correlation for few-shot segmentation~\cite{gs26_wen2025object}, explanatory instructions for generalizing across vision tasks~\cite{gs26_shen2025explanatory}, global-local alignment for 3D scene completion~\cite{gs26_yang2026gala}, low-light video question answering~\cite{gs26_yang2025vqals}, and image-text alignment for hyperspectral change detection~\cite{gs26_wu2026contrastive}. These works underscore the value of local visual evidence, correspondence, and structural organization.

Unlike prior within-source jigsaw formulations, GO-SSL mixes local elements from two sources and jointly recovers their source membership and within-source order. This produces verifiable supervision that directly couples cross-instance discrimination with spatial, geometric, or temporal structure recovery. Varying the source pairs also creates comparative contexts with different levels of ambiguity from the same unlabeled visual pool.

\section{Methodology}
\label{sec:method}

\begin{figure}[t]
  \centering
  \includegraphics[width=\linewidth]{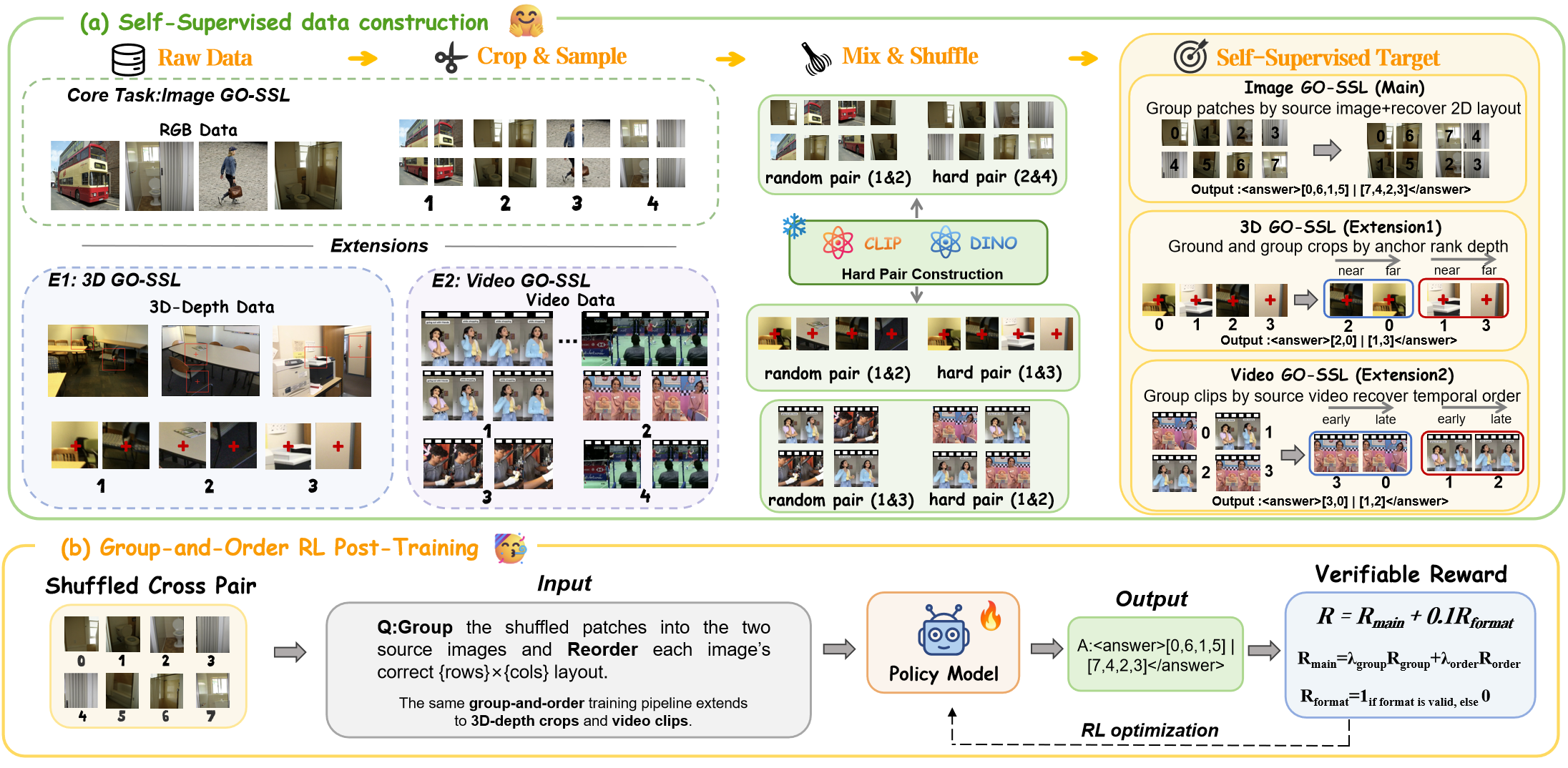}
  \caption{
  \small Overview of GO-SSL.
  \textbf{(a)} Self-supervised task construction: local elements from paired images, RGB-D scenes, or videos are shuffled and assigned verifiable group-and-order targets.
  \textbf{(b)} RL post-training: the MLLM predicts source groups and their internal spatial, depth, or temporal order, and a deterministic verifier provides rewards.
  }
  \label{fig:framework}
\end{figure}

As shown in Fig.~\ref{fig:framework}, {G}roup-and-{O}rder {S}elf-{S}upervised Reinforcement {L}earning (GO-SSL) constructs verifiable cross-instance jigsaw tasks by mixing local elements from two visual instances and requiring MLLMs to recover both instance grouping and within-instance ordering. We instantiate GO-SSL as cross-image jigsaw, cross-scene depth ranking, and cross-video temporal ordering.

\subsection{Group-and-Order Task Formulation}
\label{sec:go_formulation}
GO-SSL converts visual data into shuffled elements and asks the MLLM to recover their latent source structure. Given two visual sources \(x^A\) and \(x^B\), we extract two element sets $\mathcal{E}^A=\{e^A_1,\dots,e^A_K\}$, and $\mathcal{E}^B=\{e^B_1,\dots,e^B_K\}$, where \(\mathcal{E}^u\) denotes the \(K\) elements from source \(x^u\), and each element can be an image patch, a local 3D crop, or a video clip. Each source also has an intrinsic ordered sequence $\bar{s}^A=(e^A_1,\dots,e^A_K)$, and $\bar{s}^B=(e^B_1,\dots,e^B_K)$, corresponding to spatial raster order for images, depth order from closest to farthest for 3D, and chronological order for videos.

We mix all \(2K\) elements and apply a random permutation to obtain a shuffled sequence $\mathcal{Z}=(z_0,z_1,\dots,z_{2K-1})$, where \(z_j\) denotes the element at shuffled index \(j\). For each source \(u\in\{A,B\}\), let \(s^u=(j_1^u,\ldots,j_K^u)\) be its ground-truth ordered sequence of shuffled indices. Here, \(j_i^u\) is the shuffled index of its \(i\)-th element, so \(z_{j_i^u}=e_i^u\). Thus, \(s^u\) restores the intrinsic order of source \(x^u\) using indices in \(\mathcal{Z}\).

The model is asked to output two ordered lists of shuffled indices:
\begin{equation*}
\hat{y}=[\hat{s}_1\mid\hat{s}_2],
\end{equation*}
\begin{equation*}
\hat{s}_1,\hat{s}_2\in\{0,\ldots,2K-1\}^{K}.
\end{equation*}
Here, \(\hat{y}\) is the model prediction, \(\hat{s}_1\) and \(\hat{s}_2\) are the two predicted source groups, and \([\hat{s}_1\mid\hat{s}_2]\) denotes concatenation with a group delimiter. A valid output must use each shuffled index exactly once:
\begin{equation*}
\mathrm{set}(\hat{s}_1)\cap\mathrm{set}(\hat{s}_2)=\emptyset,
\end{equation*}
\begin{equation*}
\mathrm{set}(\hat{s}_1)\cup\mathrm{set}(\hat{s}_2)
=\{0,\ldots,2K-1\}.
\end{equation*}
For image and video GO-SSL, the two output lists are unordered at the group level, since the first predicted list may correspond to either source \(A\) or source \(B\).
We therefore evaluate both source assignments during reward computation and keep the better one, making the reward invariant to arbitrary group naming while still requiring source recovery and within-group ordering.
In contrast, 3D GO-SSL uses explicit anchors to define group identities, so group-level swapping is not applied.

We use \emph{answer-only} outputs, requiring the model
to directly predict two ordered index lists inside
\texttt{<answer></answer>} tags.
These compact, permutation-like targets can be
deterministically verified.
Appendix~\ref{app:answer_only} provides a comparison
between direct-answer and chain-of-thought prompting.

\paragraph{Supervision multiplicity}
We quantify the source-level contexts available from a fixed pool $\mathcal{D}=\{x_i\}_{i=1}^{n}$. For a pretext constructor $c$, let $\rho_c$ be the number of distinct sources in a training context. The source-context multiplicity before stochastic transformations is
\begin{equation}
\mathrm{SCM}(\mathcal{C};n)=\sum_{c\in\mathcal{C}}\binom{n}{\rho_c}.
\end{equation}
Single-source jigsaw supplies $n$ contexts (or approximately $T\cdot n$ for $T$ task types), whereas pairwise GO-SSL permits $\binom{n}{2}=O(n^2)$ source pairs. We sample a finite subset with the random-to-hard curriculum; shuffling and patch granularity add further target diversity. Training recipes are compared in Appendix~\ref{app:training_recipe_comparison}.

The combinatorial count is a property of the task construction, not a measured sample-complexity or compute-efficiency guarantee; a fixed training budget uses only a subset of the available pairs.

\subsection{Task Instantiations}

The formulation above is modality-agnostic. The three GO-SSL tasks share the same general formulation,
but differ in their source elements, within-source
ordering criteria, and group-assignment rules. For each source $u\in\{A,B\}$, let
$\mathcal{I}_u$ denote its shuffled index set.
We use $\operatorname{argsort}_{j\in\mathcal{I}_u}$
to denote the sequence of shuffled indices sorted
in ascending order by the specified criterion.

\paragraph{Cross-image group-and-order}
Our primary task is cross-image jigsaw. Given two images \(x^A\) and \(x^B\), we first center-crop
each image into a square and divide it into an \(m\times n\) grid, yielding \(K=m\cdot n\) patches per
image. The patch order follows row-major order, from left to right and top to bottom. Let \(p^u_i\)
denote the \(i\)-th patch of image \(x^u\) in this order. After the \(2K\) patches from both images
are shuffled into \(\mathcal{Z}\), the ground-truth target is
\begin{equation*}
y_{\mathrm{img}}
=[s^A_{\mathrm{img}}\mid s^B_{\mathrm{img}}],\\
\end{equation*}
\begin{equation*}
s_{\mathrm{img}}^u
=
\operatorname{argsort}^{\mathrm{lex}}_{j\in\mathcal{I}_u}
\bigl(r(z_j),c(z_j)\bigr).
\end{equation*}
Here, $r(z_j)$ and $c(z_j)$ denote the original row
and column positions of patch $z_j$, respectively.
Lexicographic sorting recovers the row-major order
of each source image, yielding $s_{\mathrm{img}}^u$.
The two groups in $y_{\mathrm{img}}$ are evaluated
up to group-level swap invariance, as defined in
the general formulation.

Unlike single-image jigsaw, ours requires the model to separate visually mixed sources before reconstructing each layout, encouraging cross-image discrimination beyond local continuity.

\paragraph{3D depth group-and-order}
We extend GO-SSL to RGB-D scenes. Each sample contains two anchor images \(x^A\) and \(x^B\),
and \(2K\) shuffled local crops. For each anchor \(x^u\), we select \(K\) marked regions with valid
depth values and crop local patches around them. Let \(c^u_i\) denote the \(i\)-th crop after sorting
the crops of anchor \(x^u\) from closest to farthest, \ie, $d(c^u_1)<d(c^u_2)<\cdots<d(c^u_K)$, where \(d(c^u_i)\) is the depth of the marked center point in crop \(c^u_i\). After all crops are
shuffled into \(\mathcal{Z}\), the ground-truth target is
\begin{equation*}
y_{\mathrm{3D}}
=[s^A_{\mathrm{3D}}\mid s^B_{\mathrm{3D}}],
\end{equation*}
\begin{equation*}
s_{\mathrm{3D}}^u
=
\operatorname{argsort}_{j\in\mathcal{I}_u}
d(z_j).
\end{equation*}
Here, $d(z_j)$ denotes the depth of the marked
center point in crop $z_j$.
Sorting by depth yields $s_{\mathrm{3D}}^u$,
which contains crops from anchor $x^u$
ordered from closest to farthest.
The two groups in $y_{\mathrm{3D}}$ correspond
to anchors $A$ and $B$, respectively.

The model is not asked to compare absolute depth values across two different images; instead, the
task combines cross-scene crop grounding with per-scene ordinal depth reasoning.

\paragraph{Video group-and-order}
We further instantiate GO-SSL for video understanding. Given two videos \(v^A\) and \(v^B\), we
sample \(K\) clips from each video in chronological order. Let \(q^u_i\) denote the \(i\)-th clip of
video \(v^u\) in this order. After all \(2K\) clips are shuffled into \(\mathcal{Z}\), the ground-truth
target is
\begin{equation*}
y_{\mathrm{vid}}
=[s^A_{\mathrm{vid}}\mid s^B_{\mathrm{vid}}],
\end{equation*}
\begin{equation*}
s_{\mathrm{vid}}^u
=
\operatorname{argsort}_{j\in\mathcal{I}_u}
\tau(z_j).
\end{equation*}
Here, $\tau(z_j)$ denotes the original temporal
position of clip $z_j$ within its source video.
Sorting by this position yields $s_{\mathrm{vid}}^u$,
which recovers the chronological order of clips
from video $v^u$.
The two groups in $y_{\mathrm{vid}}$ are evaluated
up to group-level swap invariance.

Compared with standard video ordering, cross-video G\&O prevents the model from solving the
task only through local temporal continuity. Because clips from two different videos are mixed
together, it must simultaneously capture source-level coherence and temporal progression,
which is expected to strengthen temporal reasoning, event continuity modeling, and broader video comprehension.

\subsection{Reward Design}

GO-SSL uses deterministic rewards because source identity and within-source order are known by construction. The total reward is
\begin{equation}
r = r_{\mathrm{fmt}} + r_{\mathrm{main}},
\label{eq:total_reward}
\end{equation}
where \(r_{\mathrm{main}}\leq 1\) measures group-and-order correctness, and \(r_{\mathrm{fmt}}=0.1\) only when the output follows the required \texttt{<answer></answer>} format. Exact predictions receive full reward, while valid imperfect predictions receive discounted partial credit. The reward expressions in Fig.~\ref{fig:framework}(b)
schematically illustrate the component-level scores
rather than the final training reward.
The actual reward incorporates format scaling,
group-assignment invariance, and the validity and
partial-credit rules specified below.

\paragraph{Grouping and ordering reward}
The main reward evaluates source grouping and within-source ordering.
For image and video GO-SSL, the two predicted groups are unlabeled, so we consider both mappings from predicted groups to ground-truth sources.
Let \(m:\{1,2\}\rightarrow\{A,B\}\) be a bijection, where \(m(g)\) denotes the source assigned to predicted group \(g\).
For a mapping \(m\), the grouping score is
\begin{equation}
R_{\mathrm{grp}}^{(m)}
=
\frac{1}{2K}
\sum_{g=1}^{2}
\sum_{i=1}^{K}
\mathbf{1}\!\left[\mathrm{src}(\hat{s}_{g,i}) = m(g)\right],
\end{equation}
where \(\hat{s}_{g,i}\) is the \(i\)-th shuffled index in predicted group \(\hat{s}_g\), and \(\mathrm{src}(j)\in\{A,B\}\) is the ground-truth source identity of shuffled element \(z_j\).

The ordering score measures position accuracy within each matched source:
\begin{equation}
\mathrm{Acc}(a,b)
=
\frac{1}{K}\sum_{i=1}^{K}\mathbf{1}[a_i=b_i],
\end{equation}
\begin{equation}
R_{\mathrm{ord}}^{(m)}
=
\frac{1}{2}\sum_{g=1}^{2}
\mathrm{Acc}\!\left(\hat{s}_{g},s^{m(g)}\right),
\end{equation}
where \(s^A\) and \(s^B\) are the ground-truth ordered shuffled-index sequences, and \(a_i,b_i\) denote the \(i\)-th entries of two length-\(K\) index lists.
We combine grouping and ordering as
\begin{equation}
R^{(m)}
=
\lambda_{\mathrm{grp}}R_{\mathrm{grp}}^{(m)}
+\lambda_{\mathrm{ord}}R_{\mathrm{ord}}^{(m)},
\end{equation}
\begin{equation}
\tilde{R}
=
\max_{m\in\mathcal{M}}R^{(m)},
\label{eq:go_reward}
\end{equation}
where \(\mathcal{M}\) contains the two mappings for image/video GO-SSL and the fixed anchor mapping for 3D GO-SSL; \(\tilde{R}\) is the resulting best score.

\paragraph{Validity and partial credit}
Invalid outputs, including repeated, missing, out-of-range, or incorrectly sized indices, receive zero task reward.
Exact correctness is evaluated under the same group-swap invariance as in Eq.~\eqref{eq:go_reward}.
For valid predictions, we set \(r_{\mathrm{main}}=1\) if the prediction is exactly correct, and otherwise assign a discounted partial reward \(r_{\mathrm{main}}=\gamma\tilde{R}\), where \(\gamma\in(0,1)\).
This discount ensures that exact recovery is always preferred while still providing dense learning signals for difficult cross-instance puzzles.

\subsection{Curriculum Learning Strategy}

The difficulty of GO-SSL largely depends on how the two sources are paired. Unrelated sources make grouping relatively easy, while semantically similar sources or those sharing fine-grained visual patterns create more ambiguous and informative puzzles. We therefore adopt a random-to-hard curriculum, mainly instantiated in the image setting.

\paragraph{Random pairs}
In the first stage, two sources are sampled independently from the training set: $x^A,x^B \sim p_{\mathrm{data}}(x)$ and $x^A \neq x^B$. It provides a stable warm-up, allowing models to learn the answer format, grouping behavior, and basic ordering mechanics before facing highly ambiguous examples.

\paragraph{Hard pairs}
After the model learns the task structure, we increase the difficulty with hard visual pairs. We consider two variants. \textbf{CLIP-Hard (CH)} pairs are retrieved as nearest neighbors in CLIP feature space by
$\mathcal{N}_{\mathrm{CLIP}}(x)=\operatorname{TopK}_{x'\neq x} \mathrm{sim}\!\big(\phi_{\mathrm{CLIP}}(x),\phi_{\mathrm{CLIP}}(x')\big)$, where \(\phi_{\mathrm{CLIP}}(\cdot)\) denotes CLIP image features and \(\mathrm{sim}(\cdot,\cdot)\) is cosine similarity. These pairs are semantically similar and thus make cross-image grouping more ambiguous.

Our main variant is \textbf{CLIP-Blind (CB)} pairing,
inspired by the CLIP-blind image pairs introduced by
Tong et al.~\cite{tong2024eyes}. We first retrieve CLIP-near candidates and then re-rank them with DINO to favor pairs that are close under CLIP but relatively different under DINO:
\begin{equation}
\mathrm{score}(x,x')
=
\mathrm{sim}_{\mathrm{CLIP}}(x,x')
-
\lambda\,\mathrm{sim}_{\mathrm{DINO}}(x,x').
\label{eq:clip_blind}
\end{equation}
Here, \(\mathrm{sim}_{\mathrm{CLIP}}\) and \(\mathrm{sim}_{\mathrm{DINO}}\) denote cosine similarities computed from CLIP and DINO features, respectively. A high score indicates that two images are semantically related but differ in finer visual structure, encouraging patch-level discrimination beyond coarse semantic cues.

\paragraph{Curriculum schedule}
The curriculum gradually shifts from random to hard pairs. For higher-complexity variants, we further increase the number of elements per source, \eg, from \(2\times2\) to \(2\times3\) image patches.

\subsection{RL Optimization}
GO-SSL employs a GRPO-inspired group-relative
policy objective~\cite{shao2024deepseekmath}. For each prompt, the policy samples a group of candidate answers, each scored by our deterministic verifier. The resulting rewards are used to construct within-group relative advantages, encouraging higher-reward answers over lower-reward ones for the same visual puzzle.

Following the RL post-training setup of
Visual Jigsaw~\cite{wu2026visualjigsaw},
we omit explicit KL penalties and entropy regularization. In GO-SSL, each valid response must contain two ordered lists that use every shuffled index exactly once. Our deterministic verifier checks output validity and computes task rewards based on grouping and ordering correctness, while invalid permutations receive zero task reward. We use this no-KL/no-entropy setting throughout our GO-SSL experiments.

\section{Experiments}
\label{sec:Experiments}
Our evaluation first focuses on image-centric benchmarks,
followed by extensions to 3D and video.
Additional ablation studies examine the contributions
of grouping, ordering, and source-pair construction.

\subsection{Experimental Setup} We use Qwen2.5-VL-7B-Instruct~\cite{bai2025qwen25vl} as the base MLLM and post-train it with LoRA-based RL on our answer-only GO-SSL tasks. For image GO-SSL, training samples are constructed from the COCO 2014 training split~\cite{lin2014microsoft}, where two source images are paired, their local patches are shuffled together, and the model is rewarded for recovering both source groups and within-source spatial order. We further instantiate the same group-and-order formulation on ScanNet~\cite{dai2017scannet} for 3D depth ordering and LLaVA-Video~\cite{zhang2024videoinstruction} for temporal video ordering. To increase task difficulty, we adopt a random-to-hard curriculum, where random image pairs are followed by CLIP-Hard or CLIP-Blind pairs mined with CLIP~\cite{radford2021clip} and DINOv2~\cite{oquab2023dinov2} features.

Our evaluation primarily uses VLMEvalKit~\cite{duan2024vlmevalkit}
and follows official evaluation protocols whenever available. For fair comparison, all models are evaluated using the default settings of VLMEvalKit or the official evaluation protocols adopted in prior work. We also evaluate the released 7B checkpoints of Spatial-SSRL~\cite{liu2026spatialssrl} and Visual Jigsaw~\cite{wu2026visualjigsaw} under the same pipeline. Key hyperparameters are chosen based on small validation sweeps on held-out data. Key training settings, hardware configuration, hard-pair construction, and benchmark descriptions are provided in Appendices~\ref{app:benchmarks} and ~\ref{app:exp_details}.

\subsection{Main Results on Image Cross Jigsaw}
\label{sec:image_main_results}

We first evaluate GO-SSL on image-centric multimodal benchmarks.
Table~\ref{tab:image_main} compares the base Qwen2.5-VL model, prior self-supervised post-training baselines, and our cross-image G\&O variants.

Overall, GO-SSL achieves the strongest average performance among all compared models.
Image GO-SSL (CB-$2\times3$) reaches an average score of 60.90, improving the base model by 4.15 points and outperforming Spatial-SSRL and Visual Jigsaw by 1.86 and 1.77 points, respectively.
The simpler Image GO-SSL (CB-600) variant also achieves a strong average score of 60.85, showing that the gains are not tied to a single curriculum setting.
The improvements are broad but most visible on representative perception and spatial benchmarks.
For example, Image GO-SSL (CB-$2\times3$) improves performance on HRBench4K/8K~\cite{wang2025hrbench}
and V*~\cite{wu2024vstar},
suggesting stronger detail-oriented visual perception.
It also brings clear gains on SpatialEval-maze~\cite{wang2024spatialeval} and DA-2K~\cite{yang2024depthanythingv2}, indicating that cross-image grouping and ordering benefit spatial and geometric reasoning.
The gains on SugarCrepe++~\cite{dumpala2024sugarcrepepp} and Hallusion~\cite{guan2024hallusionbench} further suggest that the learned visual grounding transfers beyond the training puzzle itself.

Compared with Visual Jigsaw, GO-SSL extends single-source ordering to cross-source group-and-order recovery, making cross-instance discrimination an explicit optimization target.
Unlike Spatial-SSRL, which combines heterogeneous spatial pretext tasks, GO-SSL uses a unified objective that couples source discrimination with structural recovery.

We also note that GO-SSL uses a moderate image-side RL budget.
Spatial-SSRL constructs 81,053 self-supervised QA pairs with an additional SFT warm-up, and Visual Jigsaw trains image jigsaw for 1000 steps with global batch size 256.
By contrast, our image GO-SSL uses 1000 steps with global batch size 64, while extracting richer supervision from paired sources through joint grouping and ordering targets.
More details are provided in Appendix~\ref{app:training_recipe_comparison}.

The gains also extend to Qwen2.5-VL-3B and LLaVA-OneVision-7B~\cite{gs26_li2025llavaonevision}
(Appendix~\ref{app:model_generalization}). Targeted cross-instance evaluations further test object comparison, grouping, attribute/state discrimination, and local correspondence (Appendix~\ref{app:cross_instance_eval}).

\begin{table*}[t]
\centering
\caption{
\small Image-centric benchmark results. CB-600 uses 400 random $2\times2$ steps and 600 CLIP-Blind $2\times2$ steps; CB-2$\times$3 replaces the last 200 steps with $2\times3$ patches. Avg. is the unweighted mean across all 14 benchmark metrics; bold denotes the best score.
}
\label{tab:image_main}
\setlength{\tabcolsep}{1.5pt}

\begin{threeparttable}
\makebox[\textwidth][c]{
\resizebox{\textwidth}{!}{
\renewcommand{\arraystretch}{1.35}
\begin{tabular}{lcccc|ccccc|c|cc|cc|c}
\toprule

\multirow{4}{*}[-0.5em]{\textbf{Method}}
& \multicolumn{4}{c|}{\shortstack{\textbf{Detail Underst.}\\\textbf{\& Vis. Percept.}}}
& \multicolumn{5}{c|}{\shortstack{\textbf{Spatial}\\\textbf{Underst. \& Reason.}}}
& \multicolumn{1}{c|}{\shortstack{\textbf{Compos.}\\\textbf{Underst.}}}
& \multicolumn{2}{c|}{\textbf{General}}
& \multicolumn{2}{c|}{\textbf{Extra}}
& \multirow{4}{*}[-0.4em]{\textbf{Avg.}} \\

& \rule{0pt}{2.05cm}
\rotatebox{90}{\textbf{HRBench4K}}
& \rotatebox{90}{\textbf{HRBench8K}}
& \rotatebox{90}{\textbf{V*}}
& \rotatebox{90}{\textbf{MMStar}}
& \rotatebox{90}{\textbf{VSR}}
& \rotatebox{90}{\textbf{SpatialEval}}
& \rotatebox{90}{\textbf{CVBench2D}}
& \rotatebox{90}{\textbf{DA-2K}}
& \rotatebox{90}{\textbf{OmniSpatial}}
& \rotatebox{90}{\textbf{SugarCrepe++}}
& \rotatebox{90}{\textbf{SeedBench}}
& \rotatebox{90}{\textbf{NaturalBench}}
& \rotatebox{90}{\textbf{Hallusion}}
& \rotatebox{90}{\textbf{EMMA}}
& \\
\cmidrule(lr){2-15}

& \textbf{test}
& \textbf{test}
& \textbf{test}
& \textbf{cp}
& \textbf{test}
& \textbf{maze}
& \textbf{test}
& \textbf{test}
& \textbf{test}
& \textbf{test}
& \textbf{test}
& \textbf{G-acc}
& \textbf{f-acc}
& \textbf{code}
& \\
\midrule

Qwen2.5-VL-7B
& 68.50 & 65.38 & 76.96 & 70.80
& 77.31 & 27.60 & 74.99 & 54.45 & 42.66
& 61.59
& 76.95 & 33.89
& 36.99 & 26.42
& 56.75 \\
\midrule

Spatial-SSRL~\cite{liu2026spatialssrl}
& 70.00 & 69.88 & 77.49 & 72.40
& 79.36 & 36.40 & 76.10 & 59.28 & 44.29
& 62.94
& 76.78 & 34.16
& 38.15 & 29.26
& 59.04 \\

Visual Jigsaw~\cite{wu2026visualjigsaw}
& 71.75 & 71.13 & 80.63 & 72.80
& 80.36 & 30.53 & 75.12 & 60.35 & 44.49
& 63.02
& 76.95 & 35.21
& 36.99 & 28.55
& 59.13 \\
\midrule

Image GO-SSL (CB-600)
& 71.50 & 69.88 & 81.68 & 74.80
& \textbf{80.42} & \textbf{46.67} & 76.74 & \textbf{61.70} & \textbf{45.79}
& \textbf{63.17}
& \textbf{77.14} & 35.26
& \textbf{38.73} & 28.36
& 60.85 \\

Image GO-SSL (CB-2$\times$3)
& \textbf{74.00} & \textbf{72.37} & \textbf{82.20} & \textbf{75.20}
& 79.68 & 42.60 & \textbf{76.82} & 61.36 & 45.27
& 61.99
& \textbf{77.14} & \textbf{35.63}
& 38.15 & \textbf{30.14}
& \textbf{60.90} \\

(Gain)
& \textcolor{green!50!black}{+5.50}
& \textcolor{green!50!black}{+6.99}
& \textcolor{green!50!black}{+5.24}
& \textcolor{green!50!black}{+4.40}
& \textcolor{green!50!black}{+2.37}
& \textcolor{green!50!black}{+15.00}
& \textcolor{green!50!black}{+1.83}
& \textcolor{green!50!black}{+6.91}
& \textcolor{green!50!black}{+2.61}
& \textcolor{green!50!black}{+0.40}
& \textcolor{green!50!black}{+0.19}
& \textcolor{green!50!black}{+1.74}
& \textcolor{green!50!black}{+1.16}
& \textcolor{green!50!black}{+3.72}
& \textcolor{green!50!black}{\textbf{+4.15}} \\

\bottomrule
\end{tabular}
}
}
\end{threeparttable}
\end{table*}

\subsection{Cross-Modal Extensions to 3D and Video}

Although our main focus is cross-image G\&O, the same formulation can be applied whenever visual elements have both source identity and internal structure.
We instantiate this idea on 3D RGB-D crops, where ordering is defined by depth, and on video clips, where ordering is defined by time.

Table~\ref{tab:3d_main} shows that 3D GO-SSL improves the average score by 8.04 points over the base model and outperforms both Spatial-SSRL and 3D Visual Jigsaw.
The largest gains appear on DA-2K~\cite{yang2024depthanythingv2} and BLINK~\cite{fu2024blink}, suggesting stronger relative depth perception and 3D spatial grounding.
Improvements on Omni3DBench~\cite{marsili2025vadar} and OmniSpatial~\cite{jia2025omnispatial} further indicate that the learned depth-aware grouping and ordering signal transfers beyond the direct depth-ranking task.

Table~\ref{tab:video_main} shows that Video GO-SSL improves the average score by 3.34 points over the base model and 1.07 points over Video Visual Jigsaw.
The strongest gain appears on AoTBench~\cite{xue2025arrowoftime}, which directly stresses temporal order and event progression.
Gains on TempCompass~\cite{liu2024tempcompass} and Video-TT~\cite{zhang2025videott} further suggest better temporal reasoning and broader video comprehension.

The strongest empirical evidence in this paper remains the cross-image setting, while the 3D and video experiments study the generality of the group-and-order formulation.
We keep the main text concise and provide additional 3D/video ablations, including order-only and Group+Order variants, in Appendix~\ref{app:cross_modal_ablation}, and video frame-number ablations in Appendix~\ref{app:video_nframe_ablation}.

\begin{table}[t]
\centering
\caption{\small Accuracy on 3D benchmarks. GO-SSL applies grouping and ordering to cross-scene depth.}
\label{tab:3d_main}
\setlength{\tabcolsep}{4.2pt}
\renewcommand{\arraystretch}{1.08}

\resizebox{\linewidth}{!}{
\begin{tabular}{l c c c c c c c}
\toprule
\textbf{Model}
& {\textbf{Omni3DBench}}
& {\textbf{3DSRBench}}
& {\textbf{BLINK}}
& {\makecell{\textbf{Static-}\\\textbf{EmbodiedBench}}}
& {\textbf{OmniSpatial}}
& {\textbf{DA-2K}}
& {\textbf{Avg.}}
\\
\cmidrule(lr){2-7}
& \textbf{count}
& \textbf{test}
& \textbf{rd-acc}
& \textbf{test}
& \textbf{test}
& \textbf{test}
&
\\
\midrule
Qwen2.5-VL-7B
& 17.14 & 57.42 & 77.42 & 46.50 & 42.66 & 54.45 & 49.27 \\
\midrule
Spatial-SSRL~\cite{liu2026spatialssrl}
& 20.00 & 58.21 & 83.87 & 48.00 & 44.29 & 59.28 & 52.28 \\
Visual Jigsaw(3D)~\cite{wu2026visualjigsaw}
& 5.71  & 58.13 & 82.26 & 47.00 & 45.99 & 71.56 & 51.78 \\
\midrule
3D GO-SSL
& \textbf{24.29} & \textbf{58.35} & \textbf{91.13} & \textbf{49.00} & \textbf{46.18} & \textbf{74.90} & \textbf{57.31} \\

(Gain)
& \textcolor{green!50!black}{+7.15}
& \textcolor{green!50!black}{+0.93}
& \textcolor{green!50!black}{+13.71}
& \textcolor{green!50!black}{+2.50}
& \textcolor{green!50!black}{+3.52}
& \textcolor{green!50!black}{+20.45}
& \textcolor{green!50!black}{\textbf{+8.04}} \\

\bottomrule
\end{tabular}
}
\end{table}

\begin{table*}[t]
\centering
\caption{
\small Evaluation results on video benchmarks under the 16-frame setting.
Video GO-SSL extends group-and-order learning to cross-video temporal ordering and improves temporal/video understanding over baselines.
}
\label{tab:video_main}
\vspace{0.1em}
\setlength{\tabcolsep}{3.8pt}
\renewcommand{\arraystretch}{1.08}
\small

\resizebox{\textwidth}{!}{
\begin{tabular}{l c c c c c c c c}
\toprule
\textbf{Model}
& \textbf{AoTBench}
& \textbf{TUNABench}
& \textbf{VideoMME}
& \multicolumn{2}{c}{\textbf{TempCompass}}
& \textbf{VSI-Bench}
& \textbf{Video-TT}
& \textbf{Avg.} \\
\cmidrule(lr){2-2}
\cmidrule(lr){3-3}
\cmidrule(lr){4-4}
\cmidrule(lr){5-6}
\cmidrule(lr){7-7}
\cmidrule(lr){8-8}
& \textbf{vqa}
& \textbf{test}
& \textbf{wo subs}
& \textbf{mc}
& \textbf{yorn}
& \textbf{test}
& \textbf{test}
& \\
\midrule

Qwen2.5-VL-7B
& 45.52 & 53.14 & 57.44 & 71.77 & 73.46 & 32.79 & 36.90 & 53.00\\

\midrule

Visual Jigsaw(Video)~\cite{wu2026visualjigsaw}
& 51.67 & 55.10 & 58.07 & 73.10 & 74.68 & 35.39 & 38.90 & 55.27\\

\midrule

Video GO-SSL
& \textbf{55.75} & \textbf{55.45} & \textbf{58.84} & \textbf{73.29} & \textbf{76.20} & \textbf{35.55} & \textbf{39.30} & \textbf{56.34}\\

(Gain)
& \textcolor{green!50!black}{+10.23}
& \textcolor{green!50!black}{+2.31}
& \textcolor{green!50!black}{+1.40}
& \textcolor{green!50!black}{+1.52}
& \textcolor{green!50!black}{+2.74}
& \textcolor{green!50!black}{+2.76}
& \textcolor{green!50!black}{+2.40}
& \textcolor{green!50!black}{\textbf{+3.34}}\\

\bottomrule
\end{tabular}
}
\end{table*}

\subsection{Ablation Studies}
\label{sec:ablation}

This section examines the key design choices of GO-SSL
through ablation studies. Table~\ref{tab:ablation} lists each trained variant once, with the random G\&O model used as the anchor for objective and pair-construction comparisons, and CLIP-Blind-600 for scale, granularity, and decoy comparisons. More ablation studies are provided in Appendix~\ref{app:additional_ablation}.

\begin{table*}[t]
\centering
\caption{
Ablation studies of GO-SSL. Each trained variant is listed once.
``D\&V'', ``Spatial'', ``Comp.'', ``Gen.'', and ``Ext.''
denote detail understanding and visual perception,
spatial understanding and reasoning, compositional understanding,
general multimodal benchmarks, and extra benchmarks, respectively.
Avg. is computed over the same 14 benchmark metrics
as Table~\ref{tab:image_main}. Bold indicates the best result within each axis group.
}
\label{tab:ablation}

\begingroup
\footnotesize
\setlength{\tabcolsep}{2pt}
\renewcommand{\arraystretch}{1.15}
\renewcommand{\tabularxcolumn}[1]{m{#1}}

\begin{tabularx}{\textwidth}{@{}
>{\raggedright\arraybackslash}m{0.64in}
>{\raggedright\arraybackslash}m{1.08in}
>{\raggedright\arraybackslash}X
*{6}{r}
@{}}
\toprule

\textbf{Axis}
& \textbf{Variant}
& \textbf{Training setting}
& \textbf{D\&V}
& \textbf{Spatial}
& \textbf{Comp.}
& \textbf{Gen.}
& \textbf{Ext.}
& \textbf{Avg.} \\
\midrule

\multirow[c]{3}{*}{Baseline}
& Qwen2.5-VL
& --
& 70.41 & 55.40 & 61.59 & 55.42 & 31.71 & 56.75 \\

& Spatial-SSRL
& --
& 72.44 & \textbf{59.09} & 62.94 & 55.47
& \textbf{33.71} & 59.04 \\

& Visual Jigsaw
& --
& \textbf{74.08} & 58.17 & \textbf{63.02}
& \textbf{56.08} & 32.77 & \textbf{59.13} \\
\midrule

\multirow[c]{3}{*}{Objective}
& Group only
& Two images, 2$\times$2; grouping only
& 71.08 & 55.66 & 62.12 & 55.52 & 32.07 & 57.14 \\

& Order only
& Single image, 2$\times$2; ordering only
& 71.63 & 58.17 & \textbf{62.74} & 55.61
& 33.37 & 58.43 \\

& G\&O-Random$^\dagger$
& Random pairs, 2$\times$2; group + order
& \textbf{74.88} & \textbf{59.83} & 62.56
& \textbf{56.39} & \textbf{34.34} & \textbf{60.19} \\
\midrule

\multirow[c]{4}{*}{\makecell[l]{Pair /\\curriculum}}
& CLIP-Hard-600
& 400 R + 600 CH (2$\times$2)
& 74.64 & 61.01 & 62.12 & \textbf{56.56}
& 33.58 & 60.43 \\

& CLIP-Blind-400
& 400 R + 400 CB (2$\times$2)
& 74.03 & 61.14 & 62.83 & 56.22
& \textbf{34.22} & 60.40 \\

& CLIP-Blind-600$^\ddagger$
& 400 R + 600 CB (2$\times$2)
& 74.47 & \textbf{62.26} & \textbf{63.17}
& 56.20 & 33.54 & 60.85 \\

& CLIP-Blind-2$\times$3
& 400 R + 400 CB (2$\times$2) + 200 CB (2$\times$3)
& \textbf{75.94} & 61.15 & 61.99
& 56.39 & 34.15 & \textbf{60.90} \\
\midrule

\multirow[c]{2}{*}[-9pt]{\makecell[l]{Candidate\\pool}}
& \makecell[l]{GO-SSL-Decoy-\\Explicit}
& \makecell[l]{CB candidates + 2 CB decoys;\\
explicit $R_{\mathrm{sel}}$}
& 73.98 & 60.65 & \textbf{62.83}
& \textbf{56.50} & \textbf{34.70} & 60.31 \\

& \makecell[l]{GO-SSL-Decoy-\\Implicit}
& \makecell[l]{CB candidates + 2 CB decoys;\\
implicit via $R_{\mathrm{grp}}+R_{\mathrm{ord}}$}
& \textbf{74.24} & \textbf{61.23} & 62.71
& 56.29 & 34.04 & \textbf{60.46} \\
\bottomrule

\end{tabularx}

\parbox{\textwidth}{
\scriptsize
R = random pairing; CH = CLIP-Hard; CB = CLIP-Blind.
$^\dagger$ G\&O-Random is the shared anchor for objective
and pair-construction comparisons.
$^\ddagger$ CLIP-Blind-600 is the shared anchor for curriculum
scale, granularity, and decoy comparisons.
}

\endgroup
\end{table*}

\paragraph{Objective design}
We first examine the individual contributions of
cross-instance grouping and spatial ordering.
The group-only variant mixes patches from two source
images and predicts their source membership without
within-source ordering supervision. In contrast, the
order-only variant performs single-image jigsaw
reconstruction, recovering patch order without
cross-instance grouping.
Both variants improve over the Qwen2.5-VL baseline,
showing that source discrimination and structural
ordering provide useful self-supervised signals.
The full G\&O objective achieves a higher overall average,
suggesting that the two components are complementary:
grouping encourages cross-image comparison, while
ordering strengthens local-to-global spatial reconstruction.

\paragraph{Hard-pair construction}
The next analysis investigates how paired-image
selection affects training.
A key advantage of GO-SSL's cross-instance formulation is that it naturally supports hard-pair construction: the difficulty of the grouping problem can be controlled simply by changing which two source images are paired.
Compared with random pairing, CLIP-Hard improves spatial understanding, indicating that semantically similar pairs provide a stronger signal than random pairs.
Under the same training budget, CLIP-Blind achieves better overall performance than CLIP-Hard.
This is likely because CLIP-nearest pairs are mainly similar in global semantic or language-aligned space, while their local patch-level differences may still be easy to separate.
CLIP-Blind pairs, which are close in CLIP space but separated in DINO space, create a more informative ambiguity: the two images are semantically related, but their local visual structures require careful comparison.

\paragraph{Curriculum scale and granularity}
We further compare different hard-pair training lengths and jigsaw granularities.
Increasing the CLIP-Blind stage from 400 to 600 steps improves the average performance, suggesting that the model continues to benefit from challenging cross-instance comparisons.
Using a denser $2\times3$ layout yields the best detail-understanding and visual-perception score, showing that finer granularity can strengthen local detail sensitivity.
However, it does not uniformly improve all categories, indicating a trade-off between richer structural supervision and higher optimization difficulty.

\paragraph{Decoy candidate expansion} Finally, we investigate whether adding visually
similar distractor patches can further strengthen GO-SSL.
Both decoy variants add two CLIP-Blind distractor patches to the candidate pool.
\textit{GO-SSL-Decoy-Explicit} uses an additional selection reward $R_{\mathrm{sel}}$ to identify true patches, whereas \textit{GO-SSL-Decoy-Implicit} relies only on $R_{\mathrm{grp}}$ and $R_{\mathrm{ord}}$, so distractors are penalized through grouping or ordering errors.
Compared with CLIP-Blind-600, decoy variants improve extra benchmark categories but not the overall average, suggesting that distractors may improve robustness to irrelevant candidates while increasing optimization difficulty.
Thus, the simpler CLIP-Blind G\&O formulation already captures most benefits with less reward-design complexity, cf. results in Appendix~\ref{app:decoy}.

\section{Conclusion}
\label{sec:conclusion}
In this work, we introduced GO-SSL, a self-supervised RL post-training framework that improves MLLMs via cross-instance group-and-order reasoning. By grouping visual elements from different sources and recovering their internal order, GO-SSL provides a verifiable, annotation-free signal that promotes fine-grained perception, cross-instance discrimination, and spatial understanding. Experiments on image benchmarks show consistent gains over baselines and prior self-supervised post-training methods, while extensions to 3D and video demonstrate the formulation's generality. These results suggest that effective self-supervised RL for MLLMs should design verifiable pretext tasks structurally aligned with transferable visual capabilities. Future work will improve GO-SSL's efficiency and study how the G\&O principle scales to larger MLLMs and specialized visual domains.



\clearpage



\bibliographystyle{unsrtnat}
\bibliography{go_ssl_neurips2026_references}

\clearpage

\appendix

\section{Impact Statement}
\label{app:Impact Statement}
This paper presents the development, training, and evaluation of GO-SSL, a cross-instance self-supervised reinforcement learning framework for improving the vision-centric understanding of multimodal large language models. By constructing automatically verifiable group-and-order
tasks from visual data, GO-SSL reduces reliance on
task-specific human annotations, external model judges,
and manually annotated question-answer pairs. This may make visual post-training more scalable and reproducible, especially for tasks requiring fine-grained perception, spatial organization, temporal ordering, and cross-instance comparison.

The proposed method may benefit downstream applications that rely on accurate visual understanding, such as embodied agents, robotics, assistive systems, and visual reasoning tools. At the same time, stronger vision-centric MLLMs may also be used in scenarios where incorrect predictions or overconfident visual judgments could cause practical harm. Examples include safety-critical decision support, surveillance-related applications, and real-world autonomous systems. Our work does not directly target these high-stakes
deployment settings, and all experiments are conducted
in research settings using existing datasets and benchmarks.

Since GO-SSL improves model capabilities through automatically constructed self-supervised tasks, its behavior still depends on the quality, diversity, and biases of the underlying visual data. Models trained with this framework should therefore be carefully evaluated for robustness, fairness, and domain-specific reliability before deployment. We encourage future work to study safeguards, failure modes, and responsible evaluation protocols for self-supervised RL post-training of MLLMs.



\section{Answer-Only versus Chain-of-Thought Prompting}
\label{app:answer_only}

Reasoning-based prompting and visual jigsaw optimization have been explored in prior work~\cite{wei2022chainofthought,wang2025jigsawr1}, while explicit deliberation does not necessarily improve every task~\cite{liu2025mindyourstep}.
We further compare answer-only (AO) prompting and reasoning-based prompting on several vision-centric benchmarks using Qwen2.5-VL-3B and Qwen2.5-VL-7B.
These benchmarks cover image, 3D, and video understanding, allowing us to examine whether explicit reasoning is consistently beneficial across different visual domains.
For AO evaluation, we use a direct-answer prompt, \eg, ``Answer with the option letter directly.''
For CoT evaluation, we instead ask the model to reason before answering, \eg, ``Think step by step before answering. The last line of your response should be exactly in the following format: Answer: \texttt{LETTER}, where \texttt{LETTER}
is one of the options.''
As shown in Table~\ref{tab:Quantitative_ao_cot}, enabling chain-of-thought (CoT) reasoning does not consistently improve performance over direct answer prediction. CoT leads to clear drops for both 3B and 7B models, suggesting that explicit reasoning may introduce noisy intermediate steps for perception-heavy tasks where the answer can often be inferred directly from visual evidence.

The gap is more pronounced for the smaller 3B model. This observation suggests that smaller MLLMs may not always benefit from generated reasoning traces, as their intermediate rationales can be less reliable and may amplify errors rather than correct them. These preliminary results do not imply that CoT is generally ineffective; rather, they motivate our choice of answer-only supervision for jigsaw-style SSL-RL, where the target is directly verifiable and unnecessary intermediate reasoning may introduce additional noise.

\section{Benchmark Descriptions}
\label{app:benchmarks}

\begin{table*}[t]
\centering
\caption{
Comparison of answer-only and reasoning-based evaluation prompting.
Reasoning ``$\times$'' denotes direct answer-only prompting, while Reasoning ``$\checkmark$'' denotes CoT-style prompting.
``Avg.'' is computed over the listed benchmarks.
}
\label{tab:Quantitative_ao_cot}
\vspace{0.3em}
\setlength{\tabcolsep}{4pt}
\renewcommand{\arraystretch}{1.12}
\small

\resizebox{0.78\textwidth}{!}{
\begin{tabular}{l c ccccccc}
\toprule
\multirow{2}{*}{\textbf{Models}}
& \multirow{2}{*}{\textbf{Reasoning}}
& \multicolumn{6}{c}{\textbf{Benchmark}}
& \multirow{2}{*}{\textbf{Avg.}}
\\
\cmidrule(lr){3-8}
&
& \textbf{V*}
& \textbf{VSR}
& \textbf{OmniSpatial}
& \textbf{SugarCrepe++}
& \textbf{DA-2K}
& \textbf{TUNABench}
 \\
\midrule

\multicolumn{9}{c}{\textit{\textbf{3B Models}}} \\
\midrule
Qwen2.5-VL-3B & {\Large $\times$}
& 75.39 & 77.27 & 45.08 & 61.26 & 52.03 & 50.35& 60.23\\
Qwen2.5-VL-3B & {\Large $\checkmark$}
& 67.54 & 68.34 & 40.05 & 50.49 & 54.16 & 48.11& 54.78\\

\midrule
\multicolumn{9}{c}{\textit{\textbf{7B Models}}} \\
\midrule
Qwen2.5-VL-7B & {\Large $\times$}
& 76.96 & 77.31 & 42.66 & 61.59 & 54.45 & 53.14 & 61.02\\
Qwen2.5-VL-7B & {\Large $\checkmark$}
& 79.06 & 71.53 & 41.48 & 56.02 & 54.64 & 49.23& 58.66\\

\bottomrule
\end{tabular}
}
\end{table*}

We evaluate GO-SSL on a broad set of image, 3D, and video benchmarks.
The benchmark suite covers detail understanding and visual perception, spatial reasoning, compositional understanding, general multimodal capability, hallucination robustness, depth-aware 3D reasoning, 3D spatial understanding, video understanding, and related capabilities.

\paragraph{Image side}
For image-side evaluation, we consider five complementary categories.
First, we use HRBench4K/8K~\cite{wang2025hrbench}, V*~\cite{wu2024vstar} and MMStar~\cite{chen2024we} to measure detail-understanding and visual-perception, where models need to recognize subtle visual evidence, high-resolution details, and small objects in cluttered scenes.
Second, we evaluate spatial understanding with VSR~\cite{liu2023vsr}, SpatialEval~\cite{wang2024spatialeval}, CV-Bench-2D~\cite{tong2024visioncentric}, DA-2K~\cite{yang2024depthanythingv2}, and OmniSpatial~\cite{jia2025omnispatial}, covering spatial relations, 2D layout, object counting, relative depth, and complex spatial reasoning.
Third, we use SugarCrepe++~\cite{dumpala2024sugarcrepepp} to evaluate compositional understanding, especially whether models can distinguish subtle changes in objects, attributes, and relations.
Finally, we include SEEDBench~\cite{li2024seedbench} and NaturalBench~\cite{li2024naturalbench} for general multimodal understanding, and HallusionBench~\cite{guan2024hallusionbench} and EMMA~\cite{hao2025emma} for hallucination-related robustness and broader multimodal reasoning.

\paragraph{3D side}
For 3D-side evaluation, we use Omni3DBench~\cite{marsili2025vadar}, 3DSRBench~\cite{ma20253dsrbench}, BLINK~\cite{fu2024blink}, StaticEmbodiedBench~\cite{xiao2025staticembodiedbench}, OmniSpatial~\cite{jia2025omnispatial}, and DA-2K~\cite{yang2024depthanythingv2}.
These benchmarks evaluate different aspects of 3D-aware perception and reasoning.
Omni3DBench and 3DSRBench focus on object-centric and scene-level 3D spatial understanding, while DA-2K directly tests relative depth judgment.
BLINK includes perception tasks such as relative depth estimation, visual correspondence, and multi-view reasoning, which are closely related to 3D-aware visual perception.
StaticEmbodiedBench further evaluates embodied spatial understanding from static scenes, and OmniSpatial provides a broader testbed for complex spatial reasoning.

\paragraph{Video side}
For video-side evaluation, we use AoTBench~\cite{xue2025arrowoftime}, TUNA-Bench~\cite{kong2025tuna}, Video-MME~\cite{fu2025videomme}, TempCompass~\cite{liu2024tempcompass}, VSI-Bench~\cite{yang2024thinkinginspace}, and Video-TT~\cite{zhang2025videott}.
These benchmarks cover complementary video abilities.
AoTBench and TempCompass emphasize temporal directionality, event order, and fine-grained temporal reasoning.
Video-MME and TUNA-Bench evaluate broader video question answering and temporal understanding under diverse video scenarios.
VSI-Bench focuses on spatial memory and reasoning in egocentric videos, while Video-TT evaluates robust temporal and video reasoning.

\section{Experimental Setup Details}
\label{app:exp_details}

\paragraph{Base model and training}
We use Qwen2.5-VL-7B-Instruct~\cite{bai2025qwen25vl} as the base MLLM for all main experiments. All GO-SSL models are post-trained with LoRA-based reinforcement learning. The LoRA rank is set to $r=32$ and the LoRA scaling factor is set to $\alpha=64$. We optimize the model with a GRPO-inspired group-relative RL objective using answer-only outputs, where the model is required to place the final group-and-order prediction inside \texttt{<answer></answer>} tags.

For image and 3D GO-SSL, we use a rollout group size of $G=8$ and a global batch size of 64. For video GO-SSL, due to the larger memory cost of video inputs, we use $G=4$ and a global batch size of 32. Unless otherwise specified, the reward weights are set to $\lambda_{\mathrm{grp}}=0.3$ and $\lambda_{\mathrm{ord}}=0.7$, the learning rate is set to $2\times10^{-5}$, and the partial reward discount is set to $\gamma=0.3$. These key hyperparameters are selected based on small-scale validation experiments. The main experiments are conducted on 8 NVIDIA A100 GPUs. All rewards are automatically computed from the self-supervised group-and-order targets, requiring no human annotations, external model judges, or additional supervised labels.

For the default GO-SSL task construction, we use a $2\times2$ patch grid for each image in image GO-SSL, sample 4 local crops for each anchor scene in 3D GO-SSL, and divide each video into 4 temporal clips for video GO-SSL.
These choices are made to balance task difficulty and training stability.
Empirically, overly simple tasks provide weak supervision and lead to limited downstream generalization, while overly difficult tasks make the group-and-order objective hard to learn and may introduce noisy rewards.

\paragraph{Training data}
For image GO-SSL, we construct cross-image jigsaw samples from the COCO 2014 training split~\cite{lin2014microsoft}. Each sample contains two source images. We crop and partition each source image into local patches, shuffle the patches from both sources together, and train the model to group patches by source image and recover the row-major spatial order within each group.

For 3D GO-SSL, we use RGB-D data from ScanNet~\cite{dai2017scannet}. Each sample contains two anchor images and shuffled local crops. The model first assigns each crop to its corresponding anchor image and then ranks the crops within each anchor from closest to farthest according to the depth of the marked center point. For video GO-SSL, we use LLaVA-Video~\cite{zhang2024videoinstruction}. Each sample contains clips from two source videos. The model is trained to group clips by source video and recover the chronological order within each video.

\paragraph{Curriculum learning}
We adopt a random-to-hard curriculum for image GO-SSL. In the warm-up stage, two source images are sampled randomly from the training set, which stabilizes format learning and basic group-and-order behavior. In the hard-pair stage, we introduce visually or semantically similar image pairs to make source grouping more ambiguous and encourage fine-grained discrimination.

We consider two hard-pair variants. \emph{CLIP-Hard} pairs are mined using OpenAI CLIP ViT-B/32 features by retrieving nearest neighbors in CLIP feature space. These pairs are semantically similar and make grouping harder than random pairing. \emph{CLIP-Blind} pairs are constructed by further filtering CLIP-near candidates with Facebook DINOv2-Base features. This favors pairs that are close under CLIP but differ under DINOv2, reducing coarse semantic shortcuts and encouraging patch-level discrimination. Unless otherwise specified, our main GO-SSL variants for image, 3D, and video use the same two-stage schedule: 400 steps with random source pairing followed by 600 steps with CLIP-Blind hard pairing.

\paragraph{Evaluation protocol}
We mainly use VLMEvalKit~\cite{duan2024vlmevalkit} for evaluation and follow its standardized protocols whenever a benchmark is supported. For benchmarks not directly covered by VLMEvalKit, we follow the official evaluation scripts or the evaluation code released by related work. For some benchmarks whose official evaluation uses a completion-style protocol, we use a strict direct-answer prompt to ensure unambiguous answer extraction and consistent comparison across models. For HallusionBench, we report f-acc using VLMEvalKit's built-in rule-based answer extraction and matching. For all compared models, including Qwen2.5-VL-7B-Instruct, Spatial-SSRL~\cite{liu2026spatialssrl}, Visual Jigsaw~\cite{wu2026visualjigsaw}, and our GO-SSL variants, we use the default evaluation settings of VLMEvalKit or the official repository. In the result tables, ``test'' typically denotes the default overall accuracy reported by the official evaluation protocol, unless otherwise specified. Metric names such as ``G-acc'' follow the official outputs, \eg, Group accuracy.

We download the released 7B checkpoints of Spatial-SSRL and Visual Jigsaw from Hugging Face and evaluate them under the same evaluation pipeline as our models. Detailed benchmark descriptions and metrics are provided in Appendix~\ref{app:benchmarks}.

\subsection{Comparison of Training Recipes}
\label{app:training_recipe_comparison}

We further compare the training recipes of GO-SSL with two closely related self-supervised RL post-training methods, Spatial-SSRL and Visual Jigsaw.
As summarized in Table~\ref{tab:training_recipe_comparison}, this comparison covers modality, task form, supervision multiplicity, the use of extra SFT, tunable parameters, and RL training budgets.
The comparison documents differences in the published training recipes. Because parameter tuning, rollout groups, and training data differ, it is not a strictly controlled attribution study; the objective ablations in Table~\ref{tab:ablation} provide a more direct comparison of grouping and ordering.

As defined in Sec.~\ref{sec:go_formulation}, supervision multiplicity measures how many distinct source-level verifiable contexts can be induced from \(n\) raw visual sources before stochastic transformations such as shuffling, cropping, masking, or prompt variations.
Under this view, a single-source jigsaw task provides \(n\) source-level contexts, while a multi-task single-source SSL pipeline with \(T\) pretext types provides approximately \(T\cdot n\) contexts.
In contrast, GO-SSL uses pairwise source construction and can draw from \(\binom{n}{2}\) cross-instance contexts.
This does not change the amount of raw data, but increases the density and diversity of supervision that can be extracted from the same training pool.

\begin{table*}[t]
\centering
\caption{
Comparison of self-supervised RL post-training recipes.
Supervision multiplicity follows the definition in Sec.~\ref{sec:go_formulation}.
}
\label{tab:training_recipe_comparison}
\vspace{0.3em}
\setlength{\tabcolsep}{3.0pt}
\renewcommand{\arraystretch}{1.12}
\small
\resizebox{\textwidth}{!}{

\begin{tabular}{l c c c c c c c c}
\toprule
\textbf{Method}

& \textbf{Modality}
& \textbf{Task form}
& \makecell{\textbf{Supervision}\\\textbf{multiplicity}}
& \textbf{Extra SFT}
& \textbf{Tuning params}
& \textbf{RL steps}
& \textbf{Global batch}
& \textbf{Rollout group}
\\
\midrule

Spatial-SSRL~\cite{liu2026spatialssrl}

& Image / RGB-D
& \makecell{Multi-task\\single-source SSL}
& \(\approx T\cdot n\) \((T=5)\)
& Yes
& Full
& 360
& 128
& 5
\\

\midrule

Visual Jigsaw-Image~\cite{wu2026visualjigsaw}

& Image
& \makecell{Single-source\\ordering}
& \(n\)
& No
& Full
& 1000
& 256
& 16
\\

Visual Jigsaw-3D~\cite{wu2026visualjigsaw}

& RGB-D
& \makecell{Single-source\\ordering}
& \(n\)
& No
& Full
& 800
& 128
& 16
\\

Visual Jigsaw-Video~\cite{wu2026visualjigsaw}

& Video
& \makecell{Single-source\\ordering}
& \(n\)
& No
& Full
& 1000
& 128
& 16
\\

\midrule

GO-SSL-Image

& Image
& \makecell{cross-instance\\group + order}
& \(\binom{n}{2}\)
& No
& LoRA
& 1000
& 64
& 8
\\

GO-SSL-3D

& RGB-D
& \makecell{cross-instance\\group + order}
& \(\binom{n}{2}\)
& No
& LoRA
& 1000
& 64
& 8
\\

GO-SSL-Video

& Video
& \makecell{cross-instance\\group + order}
& \(\binom{n}{2}\)
& No
& LoRA
& 1000
& 32
& 4
\\

\bottomrule
\end{tabular}
}
\end{table*}

Table~\ref{tab:training_recipe_comparison} highlights several practical differences.
Spatial-SSRL constructs self-supervised QA pairs from multiple heterogeneous tasks, including patch reordering, flip recognition, inpainting, depth ordering, and relative 3D position prediction.
While this design provides diverse supervision, it also requires an additional SFT cold-start stage and performs full-parameter tuning during RL post-training.
Visual Jigsaw adopts a simpler recipe by using single-source ordering tasks for images, 3D data, and videos, but each raw source is mainly used as an independent ordering instance, and the reported training uses full-parameter tuning with larger rollout groups and global batch sizes.

GO-SSL keeps the task family unified while changing the source interaction structure.
Instead of reconstructing the structure within a single source, GO-SSL pairs two sources, mixes their local elements, and requires the model to recover both source grouping and within-source order.
This makes cross-instance discrimination an explicit part of the verifiable reward.
As shown in Table~\ref{tab:training_recipe_comparison}, GO-SSL uses LoRA-based RL post-training, avoids extra SFT warm-up, and uses smaller global batch sizes than Visual Jigsaw, especially in the video setting.
Overall, the paired formulation allows GO-SSL to obtain rich supervision from a relatively simple and lightweight training recipe through different source pairings, shuffles, and hard-pair curricula.

\section{Generalization and Transfer Analyses}
\label{app:generalization}

\subsection{Generalization across Model Scales and Families}
\label{app:model_generalization}

The main experiments use Qwen2.5-VL-7B-Instruct.
To examine whether GO-SSL depends on a particular model scale or backbone family, we additionally evaluate it on Qwen2.5-VL-3B and LLaVA-OneVision-7B.
For consistency, the category averages follow the same 14 image-side benchmarks used in the main experiments.

\begin{table}[t]
\centering
\caption{Generalization across model scales and families. D\&V denotes detail understanding and visual perception; Comp., General, and Extra denote the corresponding benchmark categories. LLaVA-OV abbreviates LLaVA-OneVision.}
\label{tab:model_generalization}
\begingroup
\small
\setlength{\tabcolsep}{5pt}
\renewcommand{\arraystretch}{1.10}
\resizebox{0.80\linewidth}{!}{%
\begin{tabular}{@{}lcccccc@{}}
\toprule
\textbf{Model} & \textbf{D\&V} & \textbf{Spatial} & \textbf{Comp.} & \textbf{General} & \textbf{Extra} & \textbf{Avg.} \\
\midrule
\multicolumn{7}{c}{\textit{Qwen2.5-VL-3B}} \\
\midrule
Qwen2.5-VL-3B & 68.63 & 53.02 & 61.26 & 49.45 & 29.39 & 54.18 \\
Spatial-SSRL-3B & 70.08 & 55.66 & 60.94 & 50.22 & 32.33 & 56.05 \\
\textbf{GO-SSL-3B} & \textbf{70.78} & \textbf{56.37} & \textbf{61.31} & \textbf{50.69} & \textbf{32.51} & \textbf{56.62} \\
\midrule
\multicolumn{7}{c}{\textit{LLaVA-OneVision-7B}} \\
\midrule
LLaVA-OV-7B & 66.83 & 51.79 & \textbf{56.74} & 56.42 & 23.88 & 53.12 \\
\textbf{LLaVA-OV-7B + GO-SSL} & \textbf{69.65} & \textbf{52.77} & 56.60 & \textbf{56.65} & \textbf{25.01} & \textbf{54.46} \\
\bottomrule
\end{tabular}}
\endgroup
\end{table}

As shown in Table~\ref{tab:model_generalization}, on Qwen2.5-VL-3B, GO-SSL improves the base model by 2.44 points in average performance and outperforms Spatial-SSRL-3B across all five capability categories.
The improvement therefore persists when the backbone is reduced from 7B to 3B parameters.

GO-SSL also transfers to a different model family.
On LLaVA-OneVision-7B, it improves four of the five capability categories and raises the overall average from 53.12 to 54.46.
Together, these results indicate that the benefit of cross-instance group-and-order supervision is not specific to either the Qwen2.5-VL architecture or a single model scale.

\subsection{Cross-Backbone Generalization on 3D Benchmarks}
\label{app:llava_3d_generalization}

Beyond image-centric evaluation, we further investigate whether GO-SSL transfers to 3D visual understanding with a different backbone.
Specifically, we apply the same cross-scene group-and-order formulation to LLaVA-OneVision-7B and evaluate the resulting model on 3D benchmarks.

\begin{table}[t]
\centering
\caption{3D evaluation on LLaVA-OneVision-7B (LLaVA-OV-7B). GO-SSL improves all six benchmark scores, indicating cross-backbone transfer. The benchmark subsets and evaluation metrics are consistent with Table~\ref{tab:3d_main}.}
\label{tab:llava_3d_generalization}
\begingroup
\small
\setlength{\tabcolsep}{4pt}
\renewcommand{\arraystretch}{1.12}
\resizebox{0.93\linewidth}{!}{%
\begin{tabular}{@{}lcccccc@{}}
\toprule
\textbf{Model}
& \makecell{\textbf{Omni3D}\\\textbf{Bench}}
& \textbf{3DSRBench}
& \textbf{BLINK}
& \makecell{\textbf{Static-}\\\textbf{EmbodiedBench}}
& \makecell{\textbf{Omni}\\\textbf{Spatial}}
& \textbf{DA-2K} \\
\midrule
LLaVA-OV-7B & 56.00 & 57.42 & 62.10 & 36.50 & 44.36 & 55.03 \\
\textbf{LLaVA-OV-7B + GO-SSL-3D} & \textbf{58.67} & \textbf{58.81} & \textbf{67.74} & \textbf{39.50} & \textbf{45.27} & \textbf{63.15} \\
\bottomrule
\end{tabular}}
\endgroup
\end{table}

As shown in Table~\ref{tab:llava_3d_generalization}, LLaVA-OneVision-GO-SSL-3D improves the original backbone on all six 3D benchmarks.
The improvements are particularly notable on BLINK and DA-2K, which emphasize fine-grained spatial comparison and depth-aware reasoning.
These results further suggest that the effectiveness of GO-SSL originates from the general group-and-order formulation rather than from a specific backbone architecture.

\subsection{Cross-Domain Transfer}
\label{app:cross_domain_transfer}
\begin{table*}[t]
\centering
\caption{
Cross-domain transfer evaluation of GO-SSL variants.
Image GO-SSL and 3D GO-SSL are evaluated on both image-side and 3D-side benchmarks.
The results show that each variant is strongest in its own training domain, while still transferring non-trivially to the other domain.
}
\label{tab:cross_domain_transfer}
\vspace{0.2em}
\setlength{\tabcolsep}{3.5pt}
\renewcommand{\arraystretch}{1.08}
\small
\resizebox{\textwidth}{!}{
\begin{tabular}{l c c c c c c c c c}
\toprule
\textbf{Model}
& \makecell{\textbf{Training}\\\textbf{Domain}}
& \multicolumn{5}{c}{\textbf{Image-side Benchmarks}}
& \multicolumn{3}{c}{\textbf{3D-side Benchmarks}} \\
\cmidrule(lr){3-7}
\cmidrule(lr){8-10}
&
& \textbf{HRBench4K}
& \textbf{HRBench8K}
& \textbf{V*}
& \textbf{VSR}
& \textbf{SugarCrepe++}
& \textbf{DA-2K}
& \textbf{BLINK}
& \textbf{Omni3DBench} \\
\cmidrule(lr){3-10}
&
& test
& test
& test
& test
& test
& test
& rd-acc
& count
\\
\midrule

Qwen2.5-VL-7B
& --
& 68.50 & 65.38 & 76.96 & 77.31 & 61.59
& 54.45 & 77.42 & 17.14 \\

\midrule

Image GO-SSL
& \multirow{2}{*}{Image}
& 71.50 & 69.88 & 81.68 & 80.42 & 63.17
& 61.70 & 81.45 & 21.43 \\

(Gain)
&
& \textcolor{green!50!black}{+3.00}
& \textcolor{green!50!black}{+4.50}
& \textcolor{green!50!black}{+4.72}
& \textcolor{green!50!black}{+3.11}
& \textcolor{green!50!black}{+1.58}
& \textcolor{green!50!black}{+7.25}
& \textcolor{green!50!black}{+4.03}
& \textcolor{green!50!black}{+4.29} \\

\midrule

3D GO-SSL
& \multirow{2}{*}{3D}
& 69.38 & 67.88 & 80.10 & 78.95 & 62.14
& 74.90 & 91.13 & 24.29 \\

(Gain)
&
& \textcolor{green!50!black}{+0.88}
& \textcolor{green!50!black}{+2.50}
& \textcolor{green!50!black}{+3.14}
& \textcolor{green!50!black}{+1.64}
& \textcolor{green!50!black}{+0.55}
& \textcolor{green!50!black}{+20.45}
& \textcolor{green!50!black}{+13.71}
& \textcolor{green!50!black}{+7.15} \\

\bottomrule
\end{tabular}
}
\end{table*}
We further evaluate whether GO-SSL learns transferable visual skills beyond its training domain. Here, Image GO-SSL refers to our CLIP-Blind-600 variant in Table~\ref{tab:ablation}.
As shown in Table~\ref{tab:cross_domain_transfer}, Image GO-SSL and 3D GO-SSL both improve over the base model on image-side and 3D-side benchmarks.
Each variant performs best on the domain that matches its training signal: Image GO-SSL obtains stronger gains on image-side benchmarks such as HRBench, V*, and VSR, while 3D GO-SSL shows larger improvements on depth- and spatially-oriented benchmarks such as DA-2K, BLINK, and Omni3DBench.
At the same time, the off-domain results remain consistently positive, indicating that the group-and-order formulation does not merely overfit to a modality-specific puzzle.
Instead, cross-instance grouping and within-source ordering appear to learn general skills such as local evidence comparison, structural organization, and spatial grounding, which transfer across visual domains.

\section{Additional Ablation Studies}
\label{app:additional_ablation}

\begin{figure*}[t]
\centering
\includegraphics[width=0.85\textwidth]{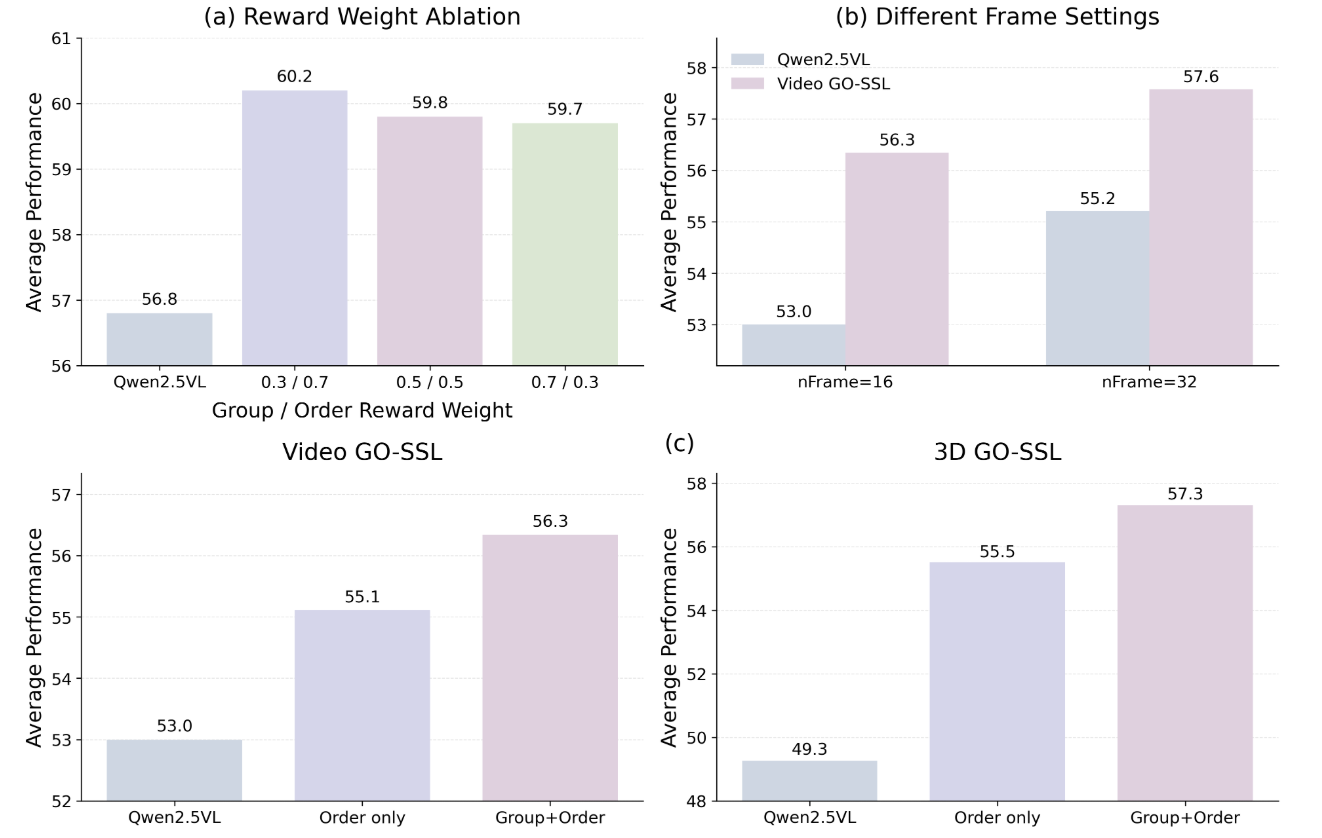}
\caption{
\small Additional ablation studies.
(a) Reward weight ablation on the image-side GO-SSL objective.
(b) Video GO-SSL under different input-frame settings.
(c) Order-only versus full Group+Order variants for video and 3D GO-SSL.
}
\label{fig:app_abl}
\end{figure*}

We provide additional ablations in Fig.~\ref{fig:app_abl} to further analyze the design choices of GO-SSL.
These studies cover three aspects: the relative weighting between grouping and ordering rewards, the robustness of Video GO-SSL under different frame budgets, and the effect of adding cross-instance grouping beyond order-only training in 3D and video domains.

\subsection{Effect of Group-and-Order Reward Weights}
\label{app:ablation_reward_weight}

GO-SSL optimizes two complementary objectives: grouping elements from the same source and recovering their within-source order.
To study their relative contribution, we vary the reward weights of the group and order terms while keeping the same training setting.
As shown in Fig.~\ref{fig:app_abl}(a), all three weight configurations clearly outperform the Qwen2.5-VL baseline, confirming that the gains do not depend on a single narrow choice of reward weighting.

To isolate the effect of reward weighting, all three variants are trained with 1000 random-pair steps and no hard-pair mining. Among the three variants, the order-heavy setting
($\lambda_{\mathrm{grp}}, \lambda_{\mathrm{ord}}) = (0.3, 0.7)$
achieves the highest average across the same 14 benchmark
metrics used in Table~1, consistent with our preliminary
validation sweep.
This suggests that spatial ordering provides a particularly strong transfer signal for image-side visual understanding, likely because recovering patch structure requires attending to local details, object parts, and their layout consistency.
The balanced and group-heavy settings remain competitive, indicating that cross-source grouping is also beneficial.
Overall, these results support our design choice of jointly optimizing grouping and ordering, with a slightly larger weight on the ordering reward.

\subsection{Cross-Modal Ablations on 3D and Video}
\label{app:cross_modal_ablation}

We further ablate whether the Group-and-Order formulation remains useful beyond static images.
For both 3D and video, we compare an order-only variant with the full Group+Order variant.
The order-only variant removes cross-instance grouping and trains the model to recover the intrinsic order within a single source: depth ordering for 3D crops and temporal ordering for video clips.
The Group+Order variant follows the full GO-SSL formulation, where elements from two sources are mixed and the model must first separate source membership and then recover the within-source order.

As shown in Fig.~\ref{fig:app_abl}(c), the order-only variants already improve over the base Qwen2.5-VL model, confirming that depth ordering and temporal ordering themselves provide useful self-supervised signals.
However, the full Group+Order variants further improve the average performance in both domains.
For video, order-only improves the average from 53.0 to 55.1, while Group+Order further improves it to 56.3.
For 3D, order-only improves the average from 49.3 to 55.5, and Group+Order further reaches 57.3.
These results support the benefit of the paired group-and-order formulation over the single-source ordering variants under our training setup.

For 3D, the improvement suggests that coupling depth ordering with cross-scene crop grounding encourages the model to compare local regions across different visual contexts, rather than only learning ordinal depth cues within one scene.
For video, the improvement indicates that cross-video grouping helps the model distinguish event-level coherence across different videos while recovering temporal order within each video.
Overall, these results support the central design of GO-SSL: grouping and ordering are complementary, and their combination provides richer verifiable supervision than ordering alone.

\subsection{Video GO-SSL under Different Frame Settings}
\label{app:video_nframe_ablation}

We also study whether Video GO-SSL remains effective under different input-frame budgets.
As shown in Fig.~\ref{fig:app_abl}(b), Video GO-SSL consistently improves over the Qwen2.5-VL baseline under both 16-frame and 32-frame evaluation settings.
With 16 frames, the average performance improves from 53.0 to 56.3.
With 32 frames, the average performance improves from 55.2 to 57.6.
This indicates that the learned temporal group-and-order ability transfers across different video sampling budgets.

The gain is larger under the 16-frame setting, where the model has fewer visual observations and therefore benefits more from improved temporal organization.
Under the 32-frame setting, the baseline becomes stronger due to richer visual input, but Video GO-SSL still provides a clear improvement.
These results suggest that Video GO-SSL does not merely exploit a specific frame configuration, but improves temporal/video understanding in a way that remains robust across input-frame settings.

\subsection{Why Does CLIP-Blind Outperform CLIP-Hard?}
\label{app:clip_blind_analysis}

We provide an intuitive explanation for why CLIP-Blind pairing leads to stronger downstream transfer than CLIP-Hard pairing in Table~\ref{tab:ablation}.
CLIP-Hard retrieves nearest neighbors in CLIP feature space, making paired images semantically similar.
However, CLIP similarity is often dominated by global language-aligned concepts, such as object category, scene type, style, or common attributes.
For a patch-level group-and-order task, such semantic similarity does not necessarily create the most informative ambiguity:
two images may be close in CLIP space while still being separable by simple local cues such as color, texture, or layout.

CLIP-Blind pairing instead keeps images close in CLIP space while penalizing high DINO similarity.
Since DINO features tend to capture visual structure and local appearance beyond high-level semantic alignment, this re-ranking favors pairs that are semantically related but differ in finer visual organization.
Such pairs encourage the model to move beyond coarse semantic cues and compare local evidence, object parts, boundaries, and spatial layouts.
Therefore, CLIP-Blind does not merely make the grouping task harder; it creates a more useful type of ambiguity, where paired images are semantically related but locally discriminative.
This better matches the downstream capabilities targeted by GO-SSL, including fine-grained perception, cross-instance comparison, and spatial understanding.
This interpretation is consistent with our results: CLIP-Hard improves spatial understanding over random pairing, while CLIP-Blind achieves stronger overall transfer.

\subsection{Analysis of Alternative Hard-Pair Constructions}
\label{app:hard_pair}

We investigate how different feature-space criteria affect
hard-pair construction and downstream transfer.
Our default CLIP-Blind strategy, referred to here as
\emph{CLIP-near + DINO-far}, favors pairs with high CLIP
similarity but relatively low DINO similarity.
As an alternative, we consider \emph{DINO-near + MAE-far},
which favors pairs that are similar in DINO feature space
but dissimilar in MAE feature space.

\begin{table}[t]
\centering
\caption{Comparison of feature-space pairing strategies under
identical 1,000-step training budgets.
CLIP-near + DINO-far corresponds to CLIP-Blind-600
in the main experiments. Differences are computed
as DINO-near + MAE-far minus CLIP-near + DINO-far.}
\label{tab:hard_pair_alt}
\begingroup
\small
\setlength{\tabcolsep}{5pt}
\renewcommand{\arraystretch}{1.12}
\resizebox{0.80\linewidth}{!}{%
\begin{tabular}{@{}lcccccc@{}}
\toprule
\textbf{Pairing Strategy}
& \textbf{D\&V}
& \textbf{Spatial}
& \textbf{Comp.}
& \textbf{Gen.}
& \textbf{Extra}
& \textbf{Avg.} \\
\midrule
CLIP-near + DINO-far
& \textbf{74.47}
& \textbf{62.26}
& \textbf{63.17}
& 56.20
& 33.54
& \textbf{60.85} \\
DINO-near + MAE-far
& 74.22
& 61.12
& 62.69
& \textbf{56.43}
& \textbf{34.38}
& 60.49 \\
\midrule
Difference
& -0.25
& -1.14
& -0.48
& +0.23
& +0.84
& -0.36 \\
\bottomrule
\end{tabular}}
\endgroup
\end{table}

As shown in Table~\ref{tab:hard_pair_alt}, both strategies
improve downstream performance but exhibit different
transfer profiles.
CLIP-near + DINO-far performs better on detail perception,
spatial reasoning, and compositional understanding,
whereas DINO-near + MAE-far achieves slightly higher
scores on the General and Extra categories.
These results suggest that different feature-space
pairing criteria provide complementary training signals.

Overall, CLIP-near + DINO-far remains our preferred
strategy, achieving the higher overall average and
stronger results on the visual capabilities most closely
related to the group-and-order objective.

\subsection{Effect of Increasing Patch Granularity}
\label{app:3x3}

The main experiments use $2\times2$ patches and progressively introduce a $2\times3$ layout.
We further examine whether GO-SSL continues to benefit from finer patch granularity by extending the final curriculum stage to $3\times3$ patches.

\begin{table}[t]
\centering
\caption{Effect of progressively finer image layouts. All variants share 400 random $2\times2$ steps followed by 400 CLIP-Blind $2\times2$ steps; only the layout of the final 200 CLIP-Blind steps differs. Avg. is computed over 14 benchmarks.}
\label{tab:grid_granularity}
\begingroup
\small
\setlength{\tabcolsep}{5pt}
\renewcommand{\arraystretch}{1.16}
\resizebox{0.90\linewidth}{!}{%
\begin{tabular}{@{}lccccccc@{}}
\toprule
\textbf{Variant} & \textbf{Final 200 steps} & \textbf{D\&V} & \textbf{Spatial} & \textbf{Comp.} & \textbf{Gen.} & \textbf{Extra} & \textbf{Avg.} \\
\midrule
CLIP-Blind-600 & $2\times2$ & 74.47 & \textbf{62.26} & \textbf{63.17} & 56.20 & 33.54 & 60.85 \\
\addlinespace[2pt]
CLIP-Blind-$2\times3$ & $2\times3$ & 75.94 & 61.15 & 61.99 & 56.39 & 34.15 & 60.90 \\
\addlinespace[2pt]
\textbf{CLIP-Blind-$3\times3$} & $3\times3$ & \textbf{76.19} & 61.11 & 61.95 & \textbf{56.53} & \textbf{34.26} & \textbf{60.99} \\
\bottomrule
\end{tabular}}
\endgroup
\end{table}

As the layout becomes finer, detail understanding and visual perception improve consistently from 74.47 to 75.94 and 76.19.
The overall average also increases from 60.85 to 60.90 and 60.99.
This suggests that finer patch granularity provides additional supervision for local visual discrimination.

The overall gains nevertheless remain modest, while spatial and compositional performance slightly decrease.
A finer layout exposes more local visual details but also introduces stronger patch-boundary fragmentation, a larger output space, and a more difficult grouping-and-ordering problem.
We therefore use the simpler $2\times2$ setting as the default and regard denser layouts as a complementary way to increase task granularity.

We also experimented with using $3\times3$ layouts from the beginning of training.
Each sample then contains $2\times9=18$ shuffled patches, compared with 8 patches for $2\times2$ and 12 patches for $2\times3$.
This substantially increases visual-token cost and combinatorial difficulty, making direct $3\times3$ training less stable under the same budget.
The progressive curriculum is therefore preferable.

\subsection{Analysis on Enlarged Candidate Pool with Decoys}
\label{app:decoy}

We further investigate whether enlarging the candidate pool with additional decoy patches can provide a stronger self-supervised signal.
Starting from the 2$\times$2 CLIP-Blind setting, we add two CLIP-Blind distractor patches to the candidate pool.
The model is still required to output two ordered groups corresponding to the two true source images, while ignoring irrelevant decoy patches.

We consider two variants.
\textit{GO-SSL-Decoy-Explicit} introduces an additional selection reward $R_{\mathrm{sel}}$ to directly supervise whether the model selects the true jigsaw patches before grouping and ordering:
\begin{equation}
R_{\mathrm{decoy}} =
\lambda_{\mathrm{sel}} R_{\mathrm{sel}}
+
\lambda_{\mathrm{grp}} R_{\mathrm{grp}}
+
\lambda_{\mathrm{ord}} R_{\mathrm{ord}} .
\end{equation}
In contrast, \textit{GO-SSL-Decoy-Implicit} removes $R_{\mathrm{sel}}$ and keeps only the original group-and-order reward:
\begin{equation}
R_{\mathrm{decoy}} =
\lambda_{\mathrm{grp}} R_{\mathrm{grp}}
+
\lambda_{\mathrm{ord}} R_{\mathrm{ord}} .
\end{equation}
In this case, selecting a decoy is penalized only implicitly through errors in grouping or ordering.

As shown in Table~\ref{tab:ablation}, both decoy variants remain competitive but do not consistently improve over the simpler CLIP-Blind curriculum.
GO-SSL-Decoy-Explicit achieves a 60.31 overall average and obtains the best external benchmark average among the decoy variants, reaching 34.70.
This suggests that explicit candidate filtering can help distractor-aware recognition.
GO-SSL-Decoy-Implicit slightly improves the overall average to 60.46, with stronger detail understanding and visual perception (74.24) and spatial understanding (61.23) than the explicit variant.
This indicates that implicit distractor rejection may better preserve the original structural reconstruction signal.

Nevertheless, both decoy variants remain below CLIP-Blind-600 and CLIP-Blind-2$\times$3 in overall average.
These results suggest that decoy-based candidate expansion is a promising but non-trivial extension.
Rather than simply increasing difficulty, it changes the task from closed-set group-and-order reconstruction to candidate selection plus reconstruction, which may partially shift optimization away from structural layout recovery.
We therefore regard it as an objective-level extension rather than a strictly controlled single-factor ablation.

\section{Targeted Evaluation of Cross-Instance Discrimination}
\label{app:cross_instance_eval}

The main benchmark suite evaluates broad visual perception, spatial reasoning, and multimodal understanding.
To more directly examine the central capability targeted by GO-SSL, we additionally evaluate models on benchmark subsets whose task definitions explicitly require comparison, grouping, correspondence, or attribute/state discrimination across visual instances.

We consider three complementary benchmark families.
VLM$^2$-Bench~\cite{gs26_zhang2025vlm2} evaluates explicit matching of visual cues across images: OC-Cpr requires cross-image object comparison, while OC-Grp requires grouping images that contain corresponding objects.
For MLLM-CompBench~\cite{gs26_kil2024compbench}, we report fine-grained visual comparison results under the VAW, State, and StateAttr metrics.
Finally, the Visual Correspondence subset of BLINK~\cite{fu2024blink} evaluates whether models can match the same local visual point across different images or viewpoints.

\begin{table}[t]
\centering
\caption{Targeted evaluation of cross-instance discrimination.
GO-SSL denotes the CB-2$\times$3 variant in
Table~\ref{tab:image_main}.
All values are accuracies (\%).}
\label{tab:cross_instance_eval}
\begingroup
\small
\setlength{\tabcolsep}{5pt}
\renewcommand{\arraystretch}{1.12}
\resizebox{0.80\linewidth}{!}{%
\begin{tabular}{@{}lcccccc@{}}
\toprule
\textbf{Model} & \textbf{OC-Cpr} & \textbf{OC-Grp} & \textbf{VAW} & \textbf{State} & \textbf{StateAttr} & \textbf{Vis. Corr.} \\
\midrule
Qwen2.5-VL-7B & 62.22 & 54.00 & 27.94 & 12.12 & 22.77 & 30.81 \\
Visual Jigsaw & 66.67 & 54.00 & 61.76 & 72.73 & 65.35 & 41.28 \\
\textbf{GO-SSL} & \textbf{71.94} & \textbf{62.00} & \textbf{75.00} & \textbf{84.85} & \textbf{78.22} & \textbf{43.60} \\
\bottomrule
\end{tabular}}
\endgroup
\end{table}

GO-SSL achieves the strongest performance on all reported metrics.
Compared with Visual Jigsaw, it improves OC-Cpr and OC-Grp by 5.27 and 8.00 points, respectively;
VAW, State, and StateAttr by 13.24, 12.12, and 12.87 points;
and Visual Correspondence by 2.32 points.

Importantly, none of these evaluations uses the Group-and-Order output format encountered during training.
The improvements therefore cannot be explained by adaptation to the training puzzle itself.
Instead, they provide more direct evidence that cross-instance group-and-order supervision transfers to the capabilities it is designed to encourage:
cross-instance object association, fine-grained comparative discrimination, source grouping, and local visual correspondence.

\section{Case Analysis and Visualization}

\subsection{Task Examples}
\label{app:task_examples}
\begin{figure}[p]
  \centering
  \includegraphics[width=0.85\linewidth]{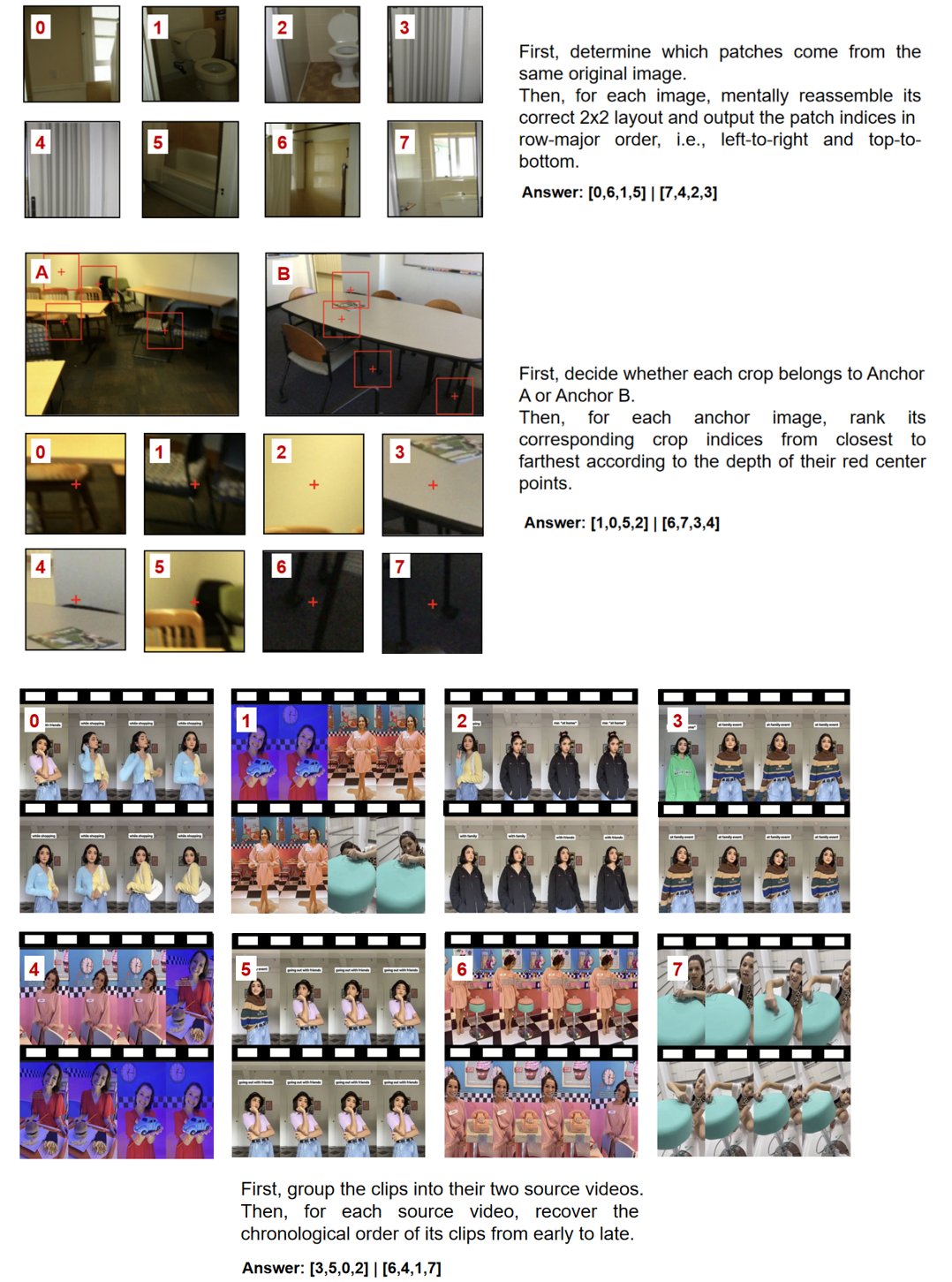}
  \caption{
    GO-SSL task examples across image, 3D, and video domains.
  }
  \label{fig:Task_Examples}
\end{figure}

Figure~\ref{fig:Task_Examples} illustrates representative GO-SSL task instances across image, 3D, and video domains.
For images and videos, the model groups shuffled elements by their source and recovers the spatial or temporal order within each group.
For 3D inputs, the model assigns local crops to their corresponding anchor scenes and orders them by the depth of the marked center points.

\subsection{Visualization Analysis}
\label{app:vis_map}
To better understand how GO-SSL affects visual grounding, we conduct an occlusion-based sensitivity analysis on binary image-text verification examples.
For each image-question pair, we ask the model whether the image matches a textual description, and estimate which regions support the predicted answer by masking local image regions.

Concretely, we divide each image into an $8 \times 8$ grid and mask one cell at a time with a gray patch.
For a target answer $a \in \{\texttt{Yes}, \texttt{No}\}$, we compute its normalized probability over the two candidate answers before and after masking cell $(r,c)$.
The occlusion sensitivity score is defined as
\begin{equation}
s_{r,c} = \max \left( p(a \mid I, q) - p(a \mid I_{\setminus (r,c)}, q), 0 \right),
\end{equation}
where $q$ is the textual query and $I_{\setminus (r,c)}$ denotes the masked image.
We normalize the resulting score map, resize it to the original resolution, and overlay it on the input image.
Unless otherwise specified, we visualize the sensitivity map with respect to each model's own predicted answer.

\begin{figure}[p]
  \centering
  \includegraphics[width=0.75\linewidth]{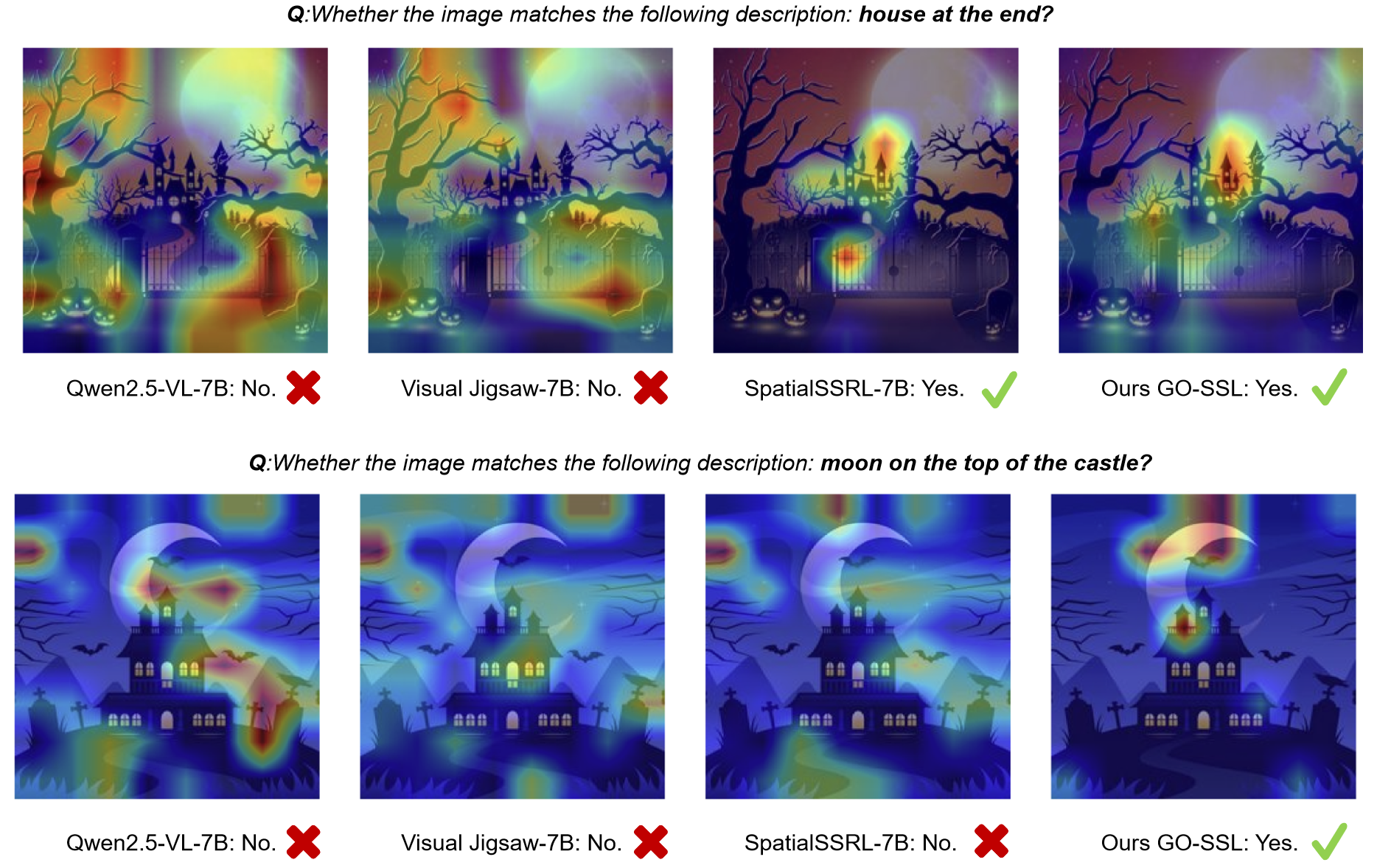}
  \caption{
    Qualitative comparison of occlusion-based sensitivity maps on fine-grained spatial matching examples.
    Compared with Qwen2.5-VL-7B and Visual Jigsaw, GO-SSL better localizes the visual evidence relevant to the queried description and produces correct answers.
    It also corrects a failure case of Spatial-SSRL in the second example, where object saliency alone is insufficient for precise spatial-relation judgment.
    }
  \label{fig:vis_map}
\end{figure}

Figure~\ref{fig:vis_map} compares Qwen2.5-VL-7B, Visual Jigsaw, Spatial-SSRL, and GO-SSL.
In the first example, which asks whether the image matches the description ``house at the end'', GO-SSL correctly focuses on the relevant house region, while the baseline models attend to broader or distracting areas.
In the second example, which requires judging whether the ``moon is on the top of the castle'', GO-SSL highlights the moon and castle roof region and predicts the correct answer.
By contrast, the other models either fail to localize the relevant evidence or make incorrect predictions despite attending to partially related regions.

These results suggest that GO-SSL encourages more localized and relation-aware visual grounding.
The grouping objective promotes discrimination between visually similar local regions across images, while the ordering objective encourages local-to-global spatial reasoning.
Together, they help the model rely more on fine-grained visual evidence rather than only global scene priors.
\subsection{Qualitative examples on different tasks}
\begin{figure}[p]
  \centering
  \includegraphics[width=0.75\linewidth]{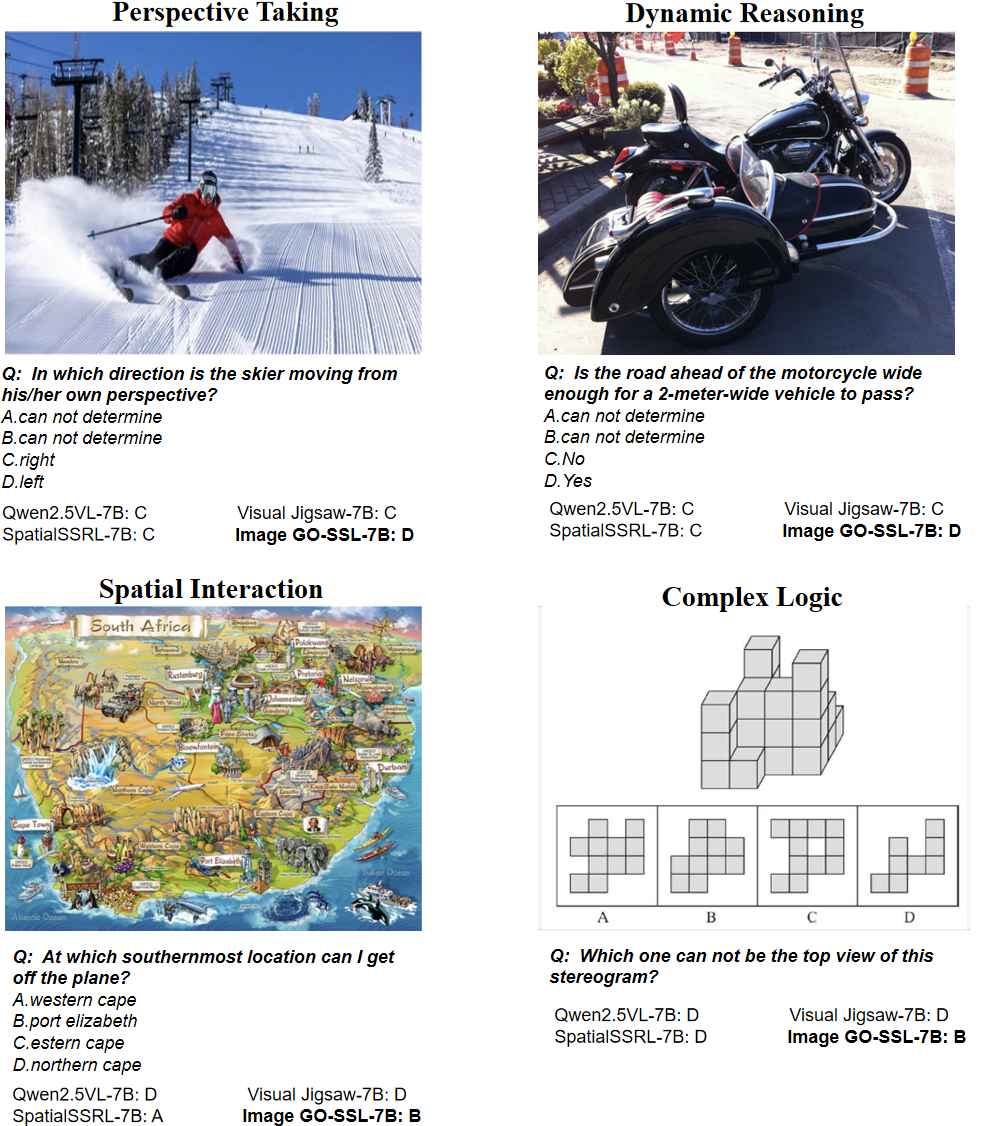}
  \caption{
    Qualitative examples on image-side tasks.
    GO-SSL gives correct predictions on perspective taking, dynamic reasoning, spatial interaction, and complex logic, where the base model and prior self-supervised post-training baselines often fail.
    }
  \label{fig:QA-image}
\end{figure}

Figure~\ref{fig:QA-image} shows qualitative examples across diverse spatial understanding and reasoning tasks, including perspective taking, dynamic reasoning, spatial interaction, and complex spatial logic.
GO-SSL gives correct predictions in these cases where the base model and prior self-supervised
post-training baselines often fail. These examples suggest that cross-instance group-and-order training
does not merely improve the model on the synthetic jigsaw objective, but also transfers to downstream
problems that require fine-grained visual grounding and relational spatial understanding.

The key intuition is that GO-SSL repeatedly forces the model to distinguish local elements from
different visual sources and then organize them into coherent structures. This encourages the model
to compare subtle visual evidence, reason over object-level relations, and reduce reliance on coarse
scene priors.

\begin{figure}[p]
  \centering
  \includegraphics[width=0.75\linewidth]{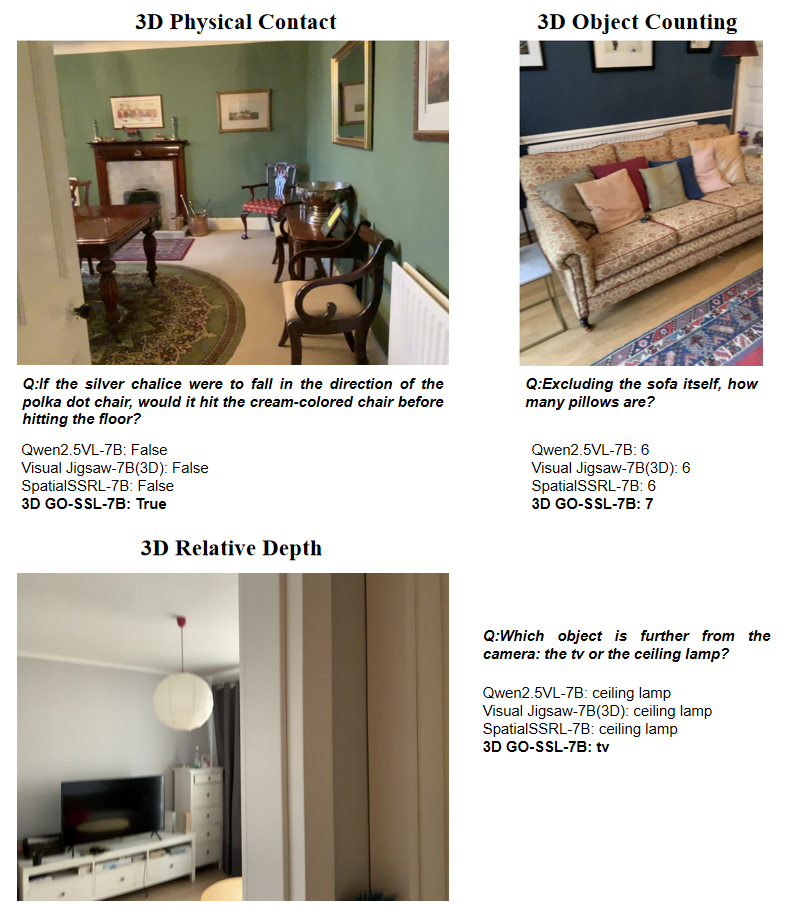}
  \caption{
    Qualitative examples on 3D-oriented visual understanding tasks.
    3D GO-SSL gives correct predictions on physical contact reasoning, object counting, and relative depth estimation, where the base model and prior self-supervised post-training baselines often fail.
    }
  \label{fig:QA-3d}
\end{figure}

\begin{figure}[p]
  \centering
  \includegraphics[width=0.99\linewidth]{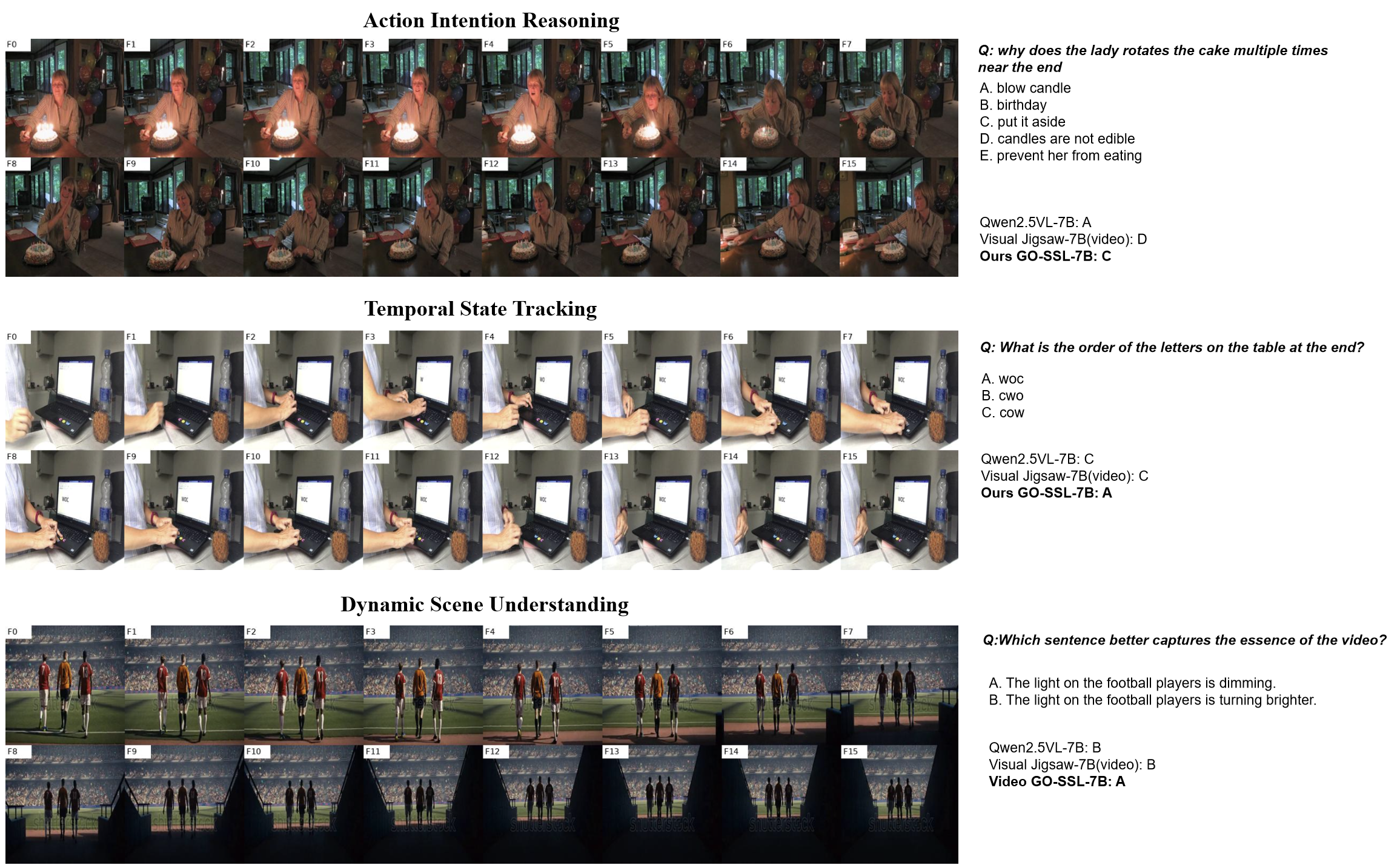}
  \caption{
Qualitative examples on video understanding tasks.
Video GO-SSL correctly answers questions requiring action intention reasoning, final-state tracking, and dynamic scene understanding, while the base model and video jigsaw baseline fail in these cases.
}
  \label{fig:QA-video}
\end{figure}

We further provide qualitative examples for the 3D and video extensions in
Figures~\ref{fig:QA-3d} and~\ref{fig:QA-video}.
For 3D-oriented tasks, 3D GO-SSL improves predictions on physical contact reasoning, object counting,
and relative depth estimation, suggesting that the learned group-and-order objective transfers to
geometric scene understanding.
For video tasks, Video GO-SSL better handles action intention, final-state tracking, and dynamic scene
changes, indicating that temporal group-and-order training can benefit downstream video understanding.

\begin{figure}[p]
    \centering
    \includegraphics[width=0.85\linewidth]{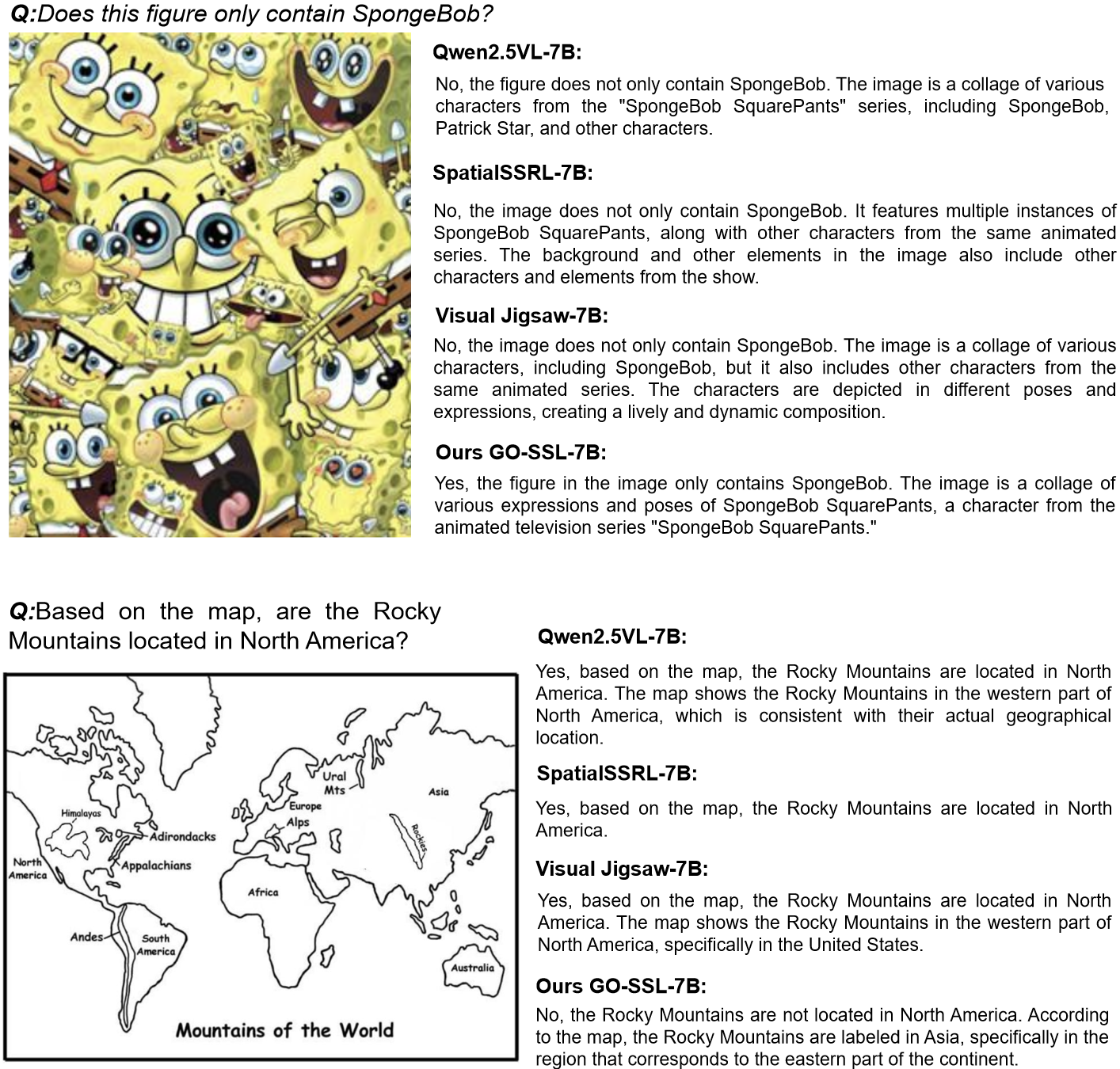}

    \vspace{0.5em}

    \includegraphics[width=0.85\linewidth]{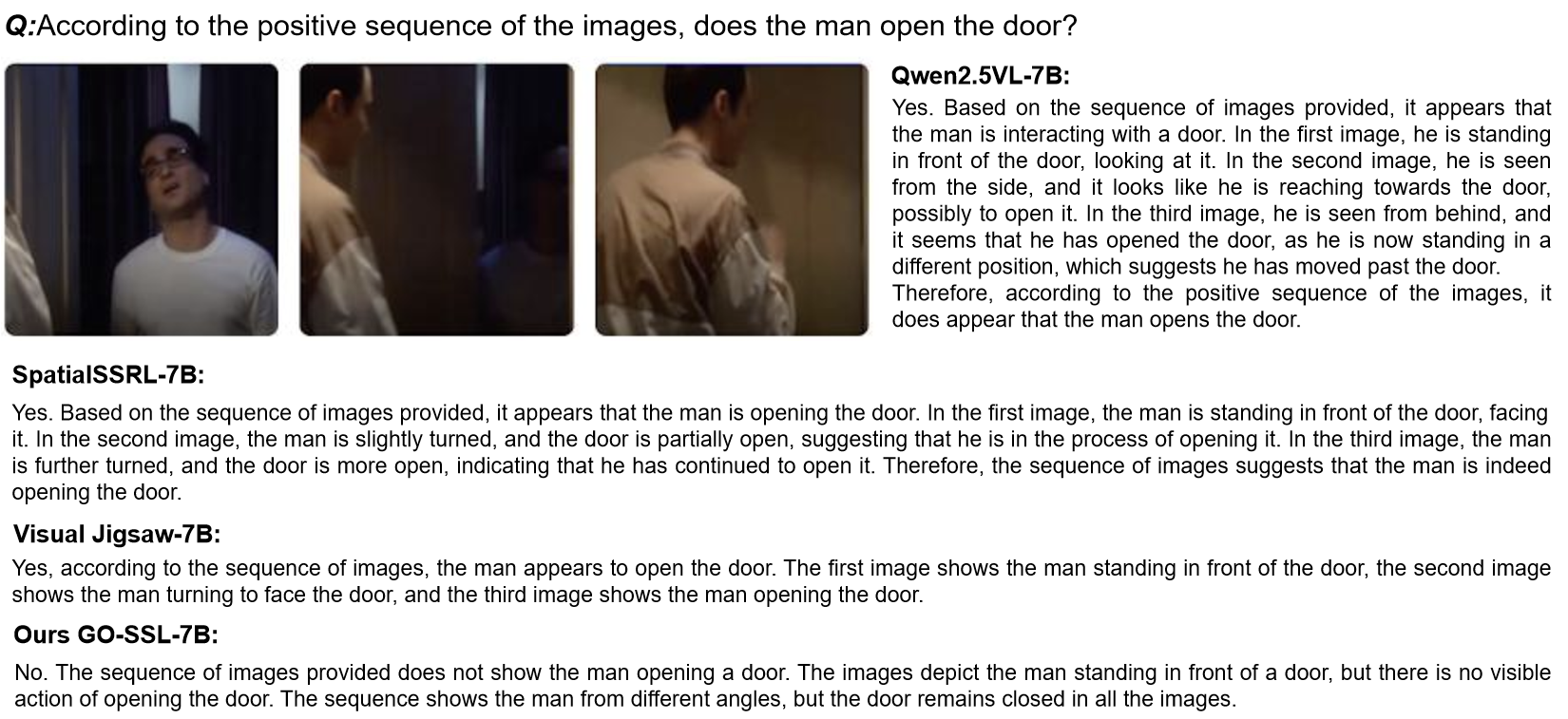}

\caption{
Qualitative examples of visually grounded reasoning across diverse scenarios.
Compared with the base model and prior self-supervised post-training baselines, GO-SSL gives more reliable answers on instance-level recognition, spatial reasoning, and video understanding, suggesting benefits beyond low-level spatial matching.
}
    \label{fig:two-cot-images}
\end{figure}

In addition, some benchmarks require not only direct visual recognition but also simple reasoning over
the visual context, such as checking whether all instances belong to the same character, reading a
map-label relation, or inferring an event from an image sequence.
As shown in Figure~\ref{fig:two-cot-images}, GO-SSL can also produce more reliable answers in such
cases, suggesting that the learned group-and-order ability provides useful inductive bias beyond low-level spatial matching.

\clearpage

\clearpage

\clearpage

\clearpage
\section{Prompt Templates}

\begin{center}
\begin{tcolorbox}[
    enhanced,
    colback=gray!10,
    colframe=black!75,
    colbacktitle=gray!18,
    coltitle=black,
    title=\textbf{Prompt for Image Cross SSL},
    fonttitle=\bfseries,
    boxrule=1.1pt,
    titlerule=0.9pt,
    arc=2.5mm,
    left=9pt,
    right=9pt,
    top=6pt,
    bottom=6pt,
    width=0.92\textwidth
]
\small
You are given n shuffled image patches that come from two \textbf{different images}. \\
Each original image was divided into a 2x2 grid, producing \textbf{k patches per image}. \\
Here are the patches, each tagged with an index reflecting the current shuffled order in which they are shown: \\

Patch 0: \texttt{<image>} \\
Patch 1: \texttt{<image>} \\
Patch 2: \texttt{<image>} \\
... \\
Patch n-1: \texttt{<image>} \\

\textbf{Task:} \\
First, determine which patches come from the same original image. \\
Then, for each image, mentally reassemble its correct 2x2 layout and output the patch indices in row-major order, \ie, left-to-right and top-to-bottom. \\

\textbf{Output format:} \\
- Output \textbf{two lists of length 4}, separated by ``|''. \\
- Each list should contain the patch indices of one source image in the correct 2x2 layout. \\
- Use each patch index exactly once across the two lists. \\
- Output \textbf{ONLY} the final answer within \texttt{<answer></answer>} tags. \\

Answer format example: \\
\texttt{<answer>[2,0,3,1] | [7,4,6,5]</answer>}
\end{tcolorbox}
\end{center}

\vspace{0.8em}

\begin{center}
\begin{tcolorbox}[
    enhanced,
    colback=gray!10,
    colframe=black!75,
    colbacktitle=gray!18,
    coltitle=black,
    title=\textbf{Prompt for 3D-Depth Cross SSL},
    fonttitle=\bfseries,
    boxrule=1.1pt,
    titlerule=0.9pt,
    arc=2.5mm,
    left=9pt,
    right=9pt,
    top=6pt,
    bottom=6pt,
    width=0.92\textwidth
]
\small
You are given two anchor images and several shuffled local crops. \\
Each anchor image contains k \textbf{red-boxed regions}. In each region, the \textbf{red center marker} indicates the exact point whose depth should be compared. \\
Each crop is taken from one marked region in one of the anchor images and contains the same red center marker. \\

Images: \\

Anchor A: \texttt{<image>} \\
Anchor B: \texttt{<image>} \\
Crop 0: \texttt{<image>} \\
Crop 1: \texttt{<image>} \\
Crop 2: \texttt{<image>} \\
... \\
Crop n-1: \texttt{<image>} \\

\textbf{Task:} \\
First, decide whether each crop belongs to Anchor A or Anchor B. \\
Then, for each anchor image, rank its corresponding crop indices from closest to farthest according to the depth of their red center points. \\
Do not rank crops by the average depth of the whole crop or the entire red-boxed region. \\
Only the depth of the red center marker should be used. \\

\textbf{Output format:} \\
- Output two lists separated by ``|''. \\
- The left list corresponds to Anchor A, and the right list corresponds to Anchor B. \\
- Each list must contain exactly k crop indices, ordered from closest to farthest. \\
- Each crop index must be used exactly once across the two lists. \\
- Output \textbf{ONLY} the final answer within \texttt{<answer></answer>} tags. \\

Answer format example: \\
\texttt{<answer>[2,5,0] | [4,1,3]</answer>}
\end{tcolorbox}
\end{center}

\vspace{0.8em}

\begin{center}
\begin{tcolorbox}[
    enhanced,
    colback=gray!10,
    colframe=black!75,
    colbacktitle=gray!18,
    coltitle=black,
    title=\textbf{Prompt for Video Cross SSL},
    fonttitle=\bfseries,
    boxrule=1.1pt,
    titlerule=0.9pt,
    arc=2.5mm,
    left=9pt,
    right=9pt,
    top=6pt,
    bottom=6pt,
    width=0.92\textwidth
]
\small
You are given n shuffled video clips that come from two \textbf{different videos}. \\
Each original video contributes \textbf{k clips}. \\

Clips in current order: \\

Clip 0: \texttt{<video>} \\
Clip 1: \texttt{<video>} \\
Clip 2: \texttt{<video>} \\
... \\
Clip n-1: \texttt{<video>} \\

\textbf{Task:} \\
First, group the clips into their two source videos. \\
Then, for each source video, recover the chronological order of its clips from early to late. \\

\textbf{Output format:} \\
- Output two lists separated by ``|''. \\
- Each list must contain exactly k clip indices. \\
- The indices refer to the current shuffled clip order and are 0-based. \\
- Each clip index must be used exactly once across the two lists. \\
- Output \textbf{ONLY} the final answer within \texttt{<answer></answer>} tags. \\

Answer format example: \\
\texttt{<answer>[2,5,1] | [0,4,3]</answer>}
\end{tcolorbox}
\end{center}


\end{document}